\documentclass[letterpaper,twocolumn,10pt]{article}
\usepackage{usenix}

\usepackage{amsmath}

\usepackage{amsthm}
\newtheorem{assumption}{Assumption}
\newtheorem{definition}{Definition}
\newtheorem{theorem}{Theorem}

\usepackage{nicefrac}
\usepackage{siunitx}
\usepackage{array,framed}
\usepackage{booktabs}
\usepackage{fix-cm}
\usepackage{
  color,
  float,
  epsfig,
  wrapfig,
  graphics,
  graphicx,
  subcaption,
  ragged2e
}

\usepackage{bm}
\usepackage{textcomp,amssymb}
\usepackage{setspace}
\usepackage{latexsym,fancyhdr,url}
\usepackage{enumerate}
\usepackage{enumitem}
\usepackage{algorithm2e}
\usepackage{algpseudocode}
\usepackage{graphics}
\usepackage{xparse} 
\usepackage{xspace}
\usepackage{multirow}
\usepackage{csvsimple}
\usepackage{balance}

\usepackage{
  tikz,
  pgfplots,
  pgfplotstable
}
\usepackage{hyperref}

\usepackage{filecontents}

\begin{document}

\date{}

\title{\Large \bf GraphToxin: Reconstructing Full Unlearned Graphs from Graph Unlearning}

\author{
{\rm Ying Song and Balaji Palanisamy}\\
University of Pittsburgh}

\maketitle

\begin{abstract}
Graph unlearning (GU) has emerged as a promising solution to comply with ``the right to be forgotten'' regulations by enabling the removal of sensitive information upon request. However, this solution is not foolproof. The involvement of multiple parties creates new attack surfaces, and residual traces of deleted data can persist within the unlearned graph neural networks (GNNs). These vulnerabilities can be exploited by attackers to recover the supposedly erased samples, undermining the intended functionality of GU. 

In this work, we propose GraphToxin, the first full graph reconstruction attack against GU. We show that GraphToxin can recover not only a deleted individual’s information and personal links but also sensitive content of their neighbors, thereby posing substantially more detrimental threats than prior membership inference attacks (MIAs). Specifically, we introduce a novel curvature matching module to provide fine-grained guidance for full unlearned graph recovery. We further extend GraphToxin to multiple-node removal under both white-box and black-box settings, showcasing its practical feasibility and potential to cause considerable harm. 

We highlight the necessity of worst-case analysis and propose a systematic evaluation framework to assess attack performance under both random and worst-case node removal scenarios. Our extensive experiments demonstrate the effectiveness and flexibility of GraphToxin. Notably, current defense mechanisms are largely ineffective against this attack. Additionally, our findings reveal that existing GU verification standards based on MIAs can be misleading: they achieve membership-level protection, while the full unlearned graph remains recoverable through GraphToxin. In light of the severe privacy risks posed by GraphToxin, our work underscores the urgent need for more robust defenses against this attack and more rigorous privacy auditing of GU. 

\end{abstract}

\section{Introduction}
\label{sec:intro} 


Graph neural networks (GNNs) have demonstrated remarkable success across diverse domains, including social network analysis, drug discovery, and financial risk management \cite{gnn_app}. Their strength lies in capturing complex structural dependencies through iterative message passing. However, this mechanism inherently amplifies the risk of sensitive information leakage, raising significant user privacy concerns. Concurrently, data protection regulations such as the General Data Protection Regulation (GDPR) \cite{gdpr} enforce the ``right to be forgotten'', requiring GNNs' capacity to remove personal data upon request. These privacy and legal imperatives have spurred the rapid emergence of graph unlearning (GU) \cite{graph_unlearn}. 

Graph unlearning aims to remove the influence of specific nodes, edges, or subgraphs from a trained GNN while preserving model performance on the remaining graph. Specifically, upon receiving an unlearning request, the model owner typically employs parameter adjustments or gradient-based optimization to mitigate these effects instead of full retraining, which is prohibitively expensive and time-consuming, and subsequently releases the unlearned version. Despite its success, GU inevitably leaves residual traces of the ``forgotten'' targets. Although such remnants are often small enough to avoid noticeable performance degradation, they can still be exploited by adversaries through membership inference \cite{unlearning_inversion_attacks_graph} or graph reconstruction attacks, thereby undermining privacy guarantees. Moreover, the involvement of multiple parties further expands the attack surface. For instance, the service providers or platforms can retain copies of the previous models prior to unlearning for legitimate reasons such as backup and debugging, while users can download such models locally on their devices. These decentralized behaviors fundamentally undermine the assurance of GU. In this work, we focus on the most challenging scenario—node unlearning \cite{surveygraphunlearning} for node-level tasks—and present the first full graph reconstruction attack against graph unlearning (GU-FGRA).

\noindent\textbf{Research Gap—Graph Unlearning} To the best of our knowledge, only few prior works \cite{unlearning_inversion_attacks_graph} have investigated the privacy risks of GU. However, their studies are limited to single-edge removal and aims to infer the link membership of a given node pair, rather than reconstructing the full unlearned graph topology. Consequently, their approach solely reveals minimal information and fails to recover confidential or sensitive content encoded in node features. Additionally, they assume that the adversary has access to partial unlearned/remaining graphs and a shadow graph with a distribution similar to that of the original graph, which are unrealistic in practice due to strict security and privacy policies. Even if partial data were leaked, finding a faithful replica of the original dataset is infeasible and using publicly available data from a different distribution can severely degrade the quality of reconstructed graphs under distributional shifts. 

\noindent\textbf{Graph Reconstruction Attack (GRA)} With respect to GRA, existing studies primarily focus on inferring link memberships \cite{link_steal_gnn, link_teller}. Only a few attempt to recover sensitive attributes \cite{mia_gra_gnn,aia_gnn} or the full graph topology \cite{inference_gnn, explain_gra}. This inconsistent terminology can be misleading, as \textit{a true full GRA should reveal all graph components}. Moreover, current research typically rely on access to node embeddings or shadow models, and often assume that sensitive node features or portions of the topology are available for recovery. They also lack comprehensive evaluation, hindering their practical applicability in scenarios such as red-teaming and model auditing. 

\noindent\textbf{Machine Unlearning (MU)} Existing reconstruction attacks against MU cannot be trivially adapted to GU-FGRA, as the complex topological structures and rich node features in graphs produce entangled patterns during GNN message passing. To date, a few works have explored reconstruction in MU. These efforts are either limited to linear models for exact unlearning \cite{bertran2024reconstruction} or recover only partial removed samples under white-box settings, with poor scalability to multiple-node removal \cite{first_unlearn_attack}. Under the i.i.d. assumption commonly adopted for tabular data, multiple sample removals can be treated as a sequence of independent single removals. In contrast, graph unlearning is fundamentally different: due to the ``ripple effects'' propagated through GNNs, even a single node removal can affect its neighbors and higher-order structures, preventing a straightforward extension to GU. 

\noindent\textbf{Research Questions} The above research gap motivates us to identify three key research questions in GU:
\begin{enumerate}
    \item Is it possible to recover full unlearned graphs under realistic white-box settings and how this attack can be extended to multiple-node removal?
    \item Can this attack be generalized to black-box settings and reconstruct full unlearned graphs using only minimal prior knowledge for multiple-node removal?
    \item How can we comprehensively evaluate the performance of this full reconstruction attack?
\end{enumerate}

\noindent\textbf{Unique Challenges and Strategies}
Beyond the challenges previously discussed, the feasibility of gradient-based GU-FGRA under realistic attack settings remains underexplored. Since no shadow graphs or models are available and only minimal prior knowledge is mastered, even a simple unlearning attack targeting a single node is an intricate task. Notably, our goal is to reveal full unlearned graphs rather than a simple combination of multiple attribute inference attacks of the removed node and multiple link membership inference attacks of its $1$-hop neighbors. This challenge is further exacerbated under multiple-node removal, where more complex structural and feature entanglements yield increasingly ambiguous gradient proxies for full graph recovery. 

To answer the key research questions and to address their unique challenges under realistic white-box settings, we present \textbf{GraphToxin, the first full graph reconstruction attack against graph unlearning}. Since the difference between publicly accessible gradients of the original and unlearned GNNs implicitly captures the information associated with the unlearned graph, GraphToxin exploits this difference for full graph recovery through three core modules. First, a gradient matching module acts as a holistic roadmap for full unlearned graph recovery by bridging the gap between the ground-truth gradient difference and the gradients generated from the recovered unlearned graph, however, it only yields a coarse proxy. Second, a novel curvature matching module provides fine-grained signals by downweighting the directions of high curvature to produce a more plausible recovery solution. Finally, a feature smoothness module enforces similarity among connected nodes as the majority of GU methods are tailored for homophilic graphs. 

GraphToxin also tackles the intractable challenge of the \textbf{data-free black-box} setting. This scenario presents a major hurdle since the adversary is restricted to using only the model's posterior probabilities instead of its gradients. This limitation inevitably introduces approximation errors, as black-box GU-FGRA must attempt to infer fine-grained graph details from coarse posterior outputs, which can degrade the reconstruction quality. To mitigate this issue, GraphToxin introduces a semantic calibration module. The goal of this module is to minimize the discrepancy between the ground-truth labels and the predictions for the recovered nodes, thereby encouraging semantically consistent reconstructions.

The effectiveness of GraphToxin is evaluated through a comprehensive evaluation framework. Existing GU methods primarily focus on small-scale random removals, which provides an incomplete assessment of reconstruction risks \cite{graph_unlearn, GIF, MEGU, ETR}. Given the inherent sparsity of real-world graphs, this operation often selects nodes that are less influential for model predictions. These ``unimportant'' nodes typically possess low degrees and cause negligible influences after removal. However, in reality, attackers would prioritize high-value targets. To better mimic actual attacks, we propose a worst-case evaluation. In our work, we define this worst-case scenario as removing the top $10\%$ of nodes with the highest degrees. Furthermore, traditional evaluation metrics are often insufficient to characterize reconstruction quality \cite{gradient_attack_gnn, graph_mia, grain}. Motivated by this limitation, we develop a systematic evaluation framework that incorporates reconstruction-related metrics to capture the global consistency of recovered features and topology, and performance-level metrics to assess functional utility of the full recovered graph. Additionally, we reveal that MIA-based GU verification standards underestimate privacy risks, even when MIA reports minimal membership-level privacy leakage, GraphToxin can still recover the full unlearned graphs. 

\noindent\textbf{Contribution} Our contributions are summarized as follows:
\begin{enumerate}[leftmargin=*]
    \item \textbf{Pioneering Attack:} We design GraphToxin, the first full graph reconstruction attack against graph unlearning. It not only recovers the removed nodes and their graph topology, but also steals the sensitive information of their neighbors, posing more detrimental privacy risks. 
    \item\textbf{Scalability and Generalization:} We successfully extend GraphToxin to multiple-node removal with comparable attack performance, and generalize GraphToxin to the black-box setting with only minimal prior knowledge.
    \item\textbf{Rigorous Evaluation Framework Design:} We propose a comprehensive evaluation framework grounded in feature-level, global-level, and performance-level perspectives. Additionally, we assess GraphToxin after removing the most influential nodes to assess its attack potency in reality. 
    \item\textbf{Empirical Validation:} The empirical results demonstrate the effectiveness and flexibility of GraphToxin across diverse graphs, GNN backbones, GU methods and white-box and black-box settings for single or multiple-node removal. We also show that existing defense methods are ineffective against GraphToxin and MIA-based safety guarantee does not imply protection from full unlearned graph recovery. This finding highlights the urgent need for developing more effective defense mechanisms against this potent attack and more rigorous privacy auditing of GU.
\end{enumerate}

\noindent\textbf{Social Impacts}
GraphToxin is significantly more detrimental than existing inference attacks, particularly in high-stake scenarios, because it recovers not only an individual's sensitive content and personal links, but also confidential information of her/his contacts. For instance, in financial crediting, revealing the full unlearned transaction graphs can expose private trades and confidential documents; in healthcare management, reconstructing full health information can fuel workplace discrimination, insurance denial, or social stigma. Once privacy is undermined, the consequences are twofold: first, it can trigger a cascade of adversarial attacks, such as re-identification or model extraction; second, organizations face liability for non-compliance with regulations and simultaneously erode users' trust in their AI systems.

\section{Background and Related Work}
\label{sec:background}

\subsection{Notations}
Given an undirected attributed graph $\mathcal{G}=(\mathcal{V}, \mathcal{E}, X)$, $\mathcal{V}$ denotes the set of nodes with $|\mathcal{V}|$ elements and each node $v$ is associated with a feature vector $X_v \in \mathbb{R}^{1 \times d}$ and a label vector $Y_v \in \mathbb{R}^{1 \times |Y|}$, where $d$ is the feature dimensionality and $|Y|$ denotes the number of labels. $\mathcal{E}$ denotes the set of edges with $|\mathcal{E}|$ elements, which can be represented by an adjacency matrix $A \in \{0,1\}^{|\mathcal{V}| \times |\mathcal{V}|}$, where $A_{w,v} = 1$ iff $(w,v) \in \mathcal{E}$, and $A_{w,v} = 0$ otherwise. For simplicity, we denote the graph as $\mathcal{G} = (A, X)$. $\mathcal{F}_{o}$, $\mathcal{F}_{u}$ and $\mathcal{F}_{a}$ indicate the original GNN, unlearned GNN, and attack model, respectively. 

\subsection{Graph Neural Networks}
The remarkable performance of most GNNs stems from the message-passing mechanism, which aggregates information from each node $v$’s local neighborhood $\mathcal{N}(v)$ and iteratively updates node embedding $H_{v}^{k}$ after $k$ layers. Formally, this process is formulated as:
\begin{equation}
    H_{v}^{k}=UPD^{k}(H_{v}^{k-1},AGG^{k-1}(\{H_{w}^{k-1}:w\in\mathcal{N}(v)\}))
\end{equation}
where $H_{v}^{0}=X_v$, and $UPD$ and $AGG$ are two arbitrary differentiable functions to design diverse GNN variants. For node classification tasks, generally, $H_{v}^{k}$ is fed into a linear classifier with a softmax function for final prediction.  

\subsection{Graph Unlearning}
We consider the most challenging GU—node unlearning throughout our paper, which is formally defined as:

\begin{definition}[\textbf{Single Node Unlearning}]
    Given the original training graph $\mathcal{G} = (A, X)$, the original trained GNN $\mathcal{F}_{o}$ and a single unlearning request $\Delta\mathcal{G}_{v}=(v, \mathcal{E}_{\mathcal{N}^1_v}, X_v)$, node unlearning aims to unlearn a single node $v$, its feature $X_v$ and its links to 1-hop neighbors $\mathcal{N}^{1}_v$ by training an unlearned model $\mathcal{F}_{u}$ on the remaining graph $\mathcal{G} \backslash \Delta\mathcal{G}_{v}$ that approximates the fully retrained model $\mathcal{F}_{o}^{\prime}(\mathcal{G} \backslash \Delta\mathcal{G}_{v})$.  
\end{definition}

\begin{definition}[\textbf{Multiple Node Unlearning}]
    Upon receiving multiple unlearning requests $\Delta\mathcal{G}=(\mathcal{V}_{\mathcal{M}}, \mathcal{N}^1_\mathcal{M}, X_\mathcal{M})$, node unlearning aims to unlearn multiple nodes $\mathcal{M}$, their node features $X_\mathcal{M}$ and links to their 1-hop neighbors $\mathcal{N}^1_\mathcal{M}$ by training unlearned model $\mathcal{F}_{u}$ on the remaining graph $\mathcal{G} \backslash \Delta\mathcal{G}$ that approximates the fully retrained model $\mathcal{F}_{o}^{\prime}(\mathcal{G} \backslash \Delta\mathcal{G})$. 
\end{definition}

Existing graph unlearning methods are commonly categorized into exact unlearning and approximate unlearning \cite{surveygraphunlearning}. GraphEraser \cite{graph_unlearn} pioneered exact GU by partitioning the training graph into disjoint shards and retraining only the shard containing the unlearning target. Wang et al. \cite{inductive_gu} further extended this framework to inductive settings. As for approximate GU, GIF \cite{GIF} leverages graph influence functions to estimate parameter changes from node deletion and applies corrective updates to approximate retraining. GNNDelete \cite{gnn_bench_nas} formulates unlearning as an optimization problem with two objectives: deleted-edge consistency and neighborhood influence preservation. MEGU \cite{MEGU} introduces a mutual evolution framework that jointly optimizes predictive and unlearning modules. ETR \cite{ETR} adopts a training-free two-stage design: an Erase stage modifies parameters to mitigate deleted-node influence, and a Rectify stage approximates gradients of remaining nodes to preserve utility.


\subsection{Graph Reconstruction Attack (GRA)}

GRAs target different graph components. They can recover partial or complete node features using released sensitive features, node embeddings or shadow graphs via attribute inference attacks \cite{mia_gra_gnn, aia_gnn}; infer the existence of specific edges using various background knowledge \cite{link_steal_gnn} or influence analysis \cite{link_teller} through link stealing attacks; or reconstruct structural information with node embeddings \cite{mia_gra_gnn} or graph encoder-decoder architectures \cite{inference_gnn} via graph topology reconstruction attacks. In this work, we focus on the most challenging setting, namely full GRA, which aims to jointly recover full node features and graph topology of the target graph.
A systematic comparison of these attack paradigms is provided in Table \ref{attack_comp}.


\begin{table*}[htbp]
\centering 
\caption{\centering \textbf{Attack Comparison}. $\surd$ represents ``access'' or ``applicable'', and $\varnothing$ denotes ``not applicable'' or ``unknown''. N/A means currently unexplored in the field.}
\label{attack_comp}
\resizebox{\linewidth}{!}{
\begin{tabular}{c|ccc|cc|cc}
\toprule
\multirow{2}{*}{\textbf{Attack Type}}         & \multicolumn{3}{c|}{\textbf{Attacker’s Knowledge and Capabilities}}                    & \multicolumn{2}{c|}{\textbf{Recovery Content}} & \multicolumn{2}{c}{\textbf{Attack Unlearning Case}}  \\ \cline{2-8}
                                              & \textbf{Graph Data}                      & \textbf{Model Access}       & \textbf{Graph Unlearning}          & \textbf{Feature}                                  & \textbf{Topology}                                & \textbf{Random}             & \textbf{Worse}               \\ \cline{1-8}
\textbf{Attribute Inference Attack (N/A)}           & Partial/Shadow Graph & Grey/Black Box       & $\varnothing$  & Single/Partial                           & $\varnothing$             & $\varnothing$ & $\varnothing$   \\
\textbf{Link Stealing Attack \cite{unlearning_inversion_attacks_graph}}                 & Partial/Shadow Graph & Grey/Black Box       & Partial Unlearned Graph      & $\varnothing$             & Single Edge                                    & $\surd$        & $\varnothing$  \\
\textbf{Graph Topology Reconstruction Attack (N/A)} & Partial/Shadow Graph & Grey/Black Box       & $\varnothing$  & $\varnothing$              & Partial/Full                            & $\varnothing$ & $\varnothing$   \\ \cline{1-8}
\textbf{Full Graph Reconstruction Attack (Ours)}     & Data-free                    & White/Black Box & \#of Node\&Edge and Labels & Full                                     & Full                                    & $\surd$        & $\surd$         \\  \bottomrule
\end{tabular}}
\end{table*}

\subsection{Problem Statement}
We formally define the full graph reconstruction attack against graph unlearning (GU-FGRA). For single-node removal, a GU-FGRA aims to recover the full unlearned graph $\mathcal{G}_{{u}_v}=(\mathcal{V}_{{u}_{v}}, \mathcal{E}_{u_v}, X_{{u}_{v}})$, where the recovered nodes $\mathcal{V}_{{u}_{v}}=[v,\mathcal{N}^{1}_{v}]$ contain the removed node $v$ and its affected 1-hop neighbors $\mathcal{N}^{1}_{v}$. Please note that $\bm{\mathcal{G}_{u_{v}}\neq\Delta\mathcal{G}_{v}}$. Unlike $\Delta\mathcal{G}_{v}$, $\mathcal{G}_{u_{v}}$ additionally recovers node features of its 1-hop neighbors. 
Moreover, recovering $\Delta\mathcal{G}_{v}$ can be simply treated as a combination of $d$ times attribute inference attacks of a single node $v$ and a graph topology reconstruction attack of its associated links $\mathcal{N}^{1}_v$, where $d$ denotes the number of node features. In contrast, recovering $\mathcal{G}_{u_{v}}$ is significantly more complicated, as it further requires the simultaneous full feature recovery of multiple recovered nodes. For multiple-node removal, attackers strive to reconstruct $\mathcal{G}_{u}=(\mathcal{V}_u, \mathcal{E}_u, X_u)$. Similarly, $\mathcal{G}_{u}\neq\Delta\mathcal{G}$ where $\mathcal{V}_u=[\mathcal{V}_{\mathcal{M}}, \mathcal{N}^1_\mathcal{M}]$ contains multiple removed nodes $\mathcal{M}$ and their 1-hop neighbors $\mathcal{N}^1_\mathcal{M}$. Under white-box settings, the adversary leverages model gradients $\nabla\mathcal{L}(\theta)$ ($\nabla\mathcal{L}$ hereinafter) to recover the full unlearned graph for both single- or multiple-node removal, while under black-box settings, she/he exploits the posterior probabilities $P=(p_1, p_2,\dots, p_{|Y|})$ obtained before and after node removals to achieve the same objective.

\section{Attack Framework Design}
In this section, we formalize the threat model and introduce GraphToxin, the first data-free GU-FGRA, followed by an evaluation framework to systematically assess it.
\label{sec:framework}

\subsection{Threat Model}

\subsubsection{\textbf{Attacker's Goal}}
The attacker aims to fully reconstruct the unlearned graph, including the removed nodes, their affected neighbors, the induced graph topology, and associated node features.  

\subsubsection{\textbf{Attacker's Knowledge and Capabilities}}

We consider a standard MLaaS deployment involving a model owner, a server, and users. The model owner trains a GNN on a private graph and uploads the GNN to the server. The server then deploys the model and packs it into an API for user queries \cite{first_unlearn_attack, unlearn_membership_attack}. Upon receiving an unlearning request from a participant (i.e., a specific node in the private training graph), the model owner adopts a GU method to remove any information associated with that participant (i.e., its node features and neighboring links), releases the unlearned GNN, and deletes the previous version in compliance with data protection regulations. However, the involvement of multiple parties and time gaps between model updates expand the attack surface, where the adversary can exploit to reconstruct the full unlearned graphs in a data-free manner. 

Our attack setting fundamentally differs from existing reconstruction attacks. They assume that the adversary either has partial access to original training or unlearned graphs, or can obtain shadow graphs \cite{unlearning_inversion_attacks_graph, link_steal_gnn, mia_gra_gnn, aia_gnn, inference_gnn}. However, these assumptions are unrealistic: first, original training graphs are proprietary and constitute valuable assets of companies and institutions, necessitating strong protections against potential attackers; second, shadow graphs typically demand same or similar distributions from the original training graphs. Since such graphs are usually protected, it is hard for the adversary to acquire the distribution information. Therefore, GU-FGRA should be strictly operated under a data-free setting. 

We consider two types of attackers: the honest-but-curious server and the malicious user. Following standard settings in gradient-based and unlearning attacks \cite{grain, realistic_gradient, gradient_attack_gnn, gia_survey, first_unlearn_attack}, the server has white-box access to model gradients of the original and unlearned GNNs, without any prior knowledge of unlearning methods, GNN architecture, and underlying distributions of original or unlearned graphs. This assumption is realistic as the server can save copies of all uploaded models or archive the deleted versions for legitimate reasons, including but not limited to backup, auditing, debugging, or rollback after accidental deletion. Meanwhile, the malicious user only has black-box access to the original and unlearned GNNs \cite{unlearn_membership_attack, unlearning_inversion_attacks_graph}. She/he can submit model queries to the server, and the server will faithfully respond to these queries and return posterior-based model predictions. This setup aligns with the standard practice across real-world MLaaS platforms, such as Microsoft Azure \cite{azure_mlflow_deployment_guidelines}, Google Vertex \cite{google_vertex_tabular_predictions} and AWS SageMaker \cite{aws_sagemaker_clarify_inference_config}; and this assumption holds as the user can locally retain historical query-response pairs or compiled inference-only model artifacts. Whenever a model update is prompted, the user adopts the same procedure to the new version without overwriting the old ones. With a chronological sequence of collected query-response pairs or model snapshots, the user can conduct data-free model stealing attacks to construct surrogate models \cite{data_free_msa_gnn}, effectively estimating gradients and capturing model shifts induced by the unlearning process. Beyond malicious actors, auditors can act as a red team to launch GU-FGRA with minimal required knowledge, enabling them to discover vulnerabilities and anomalies to improve the security and privacy of graph-based AI systems. 

We assume that the adversary has knowledge of the number of nodes and edges at each model snapshot. This assumption is practical since such information can be exposed through simple count queries, system logs, model cards \cite{model_card} and publicly available technical documents, all of which are commonly provided on MLaaS platforms to facilitate proper model usage. Such metadata also aligns with disclosure obligations under data protection regulations and AI policies such as GDPR \cite{gdpr} and EU AI Act \cite{eu_ai_act}. Moreover, consistent with standard practice \cite{first_unlearn_attack, gradient_re_full_label}, we assume the full label knowledge since attackers can employ well-established label inference techniques \cite{label_inf, image_gia,deep_leakage_gradients} to recover labels from model gradients, which serves as a cornerstone for subsequent reconstruction attacks. We further relax this assumption by introducing varying levels of label noises in Section \ref{label_noise_sec}.

\subsection{Attack Framework Design}
\begin{figure}[h]
  \centering
  \includegraphics[width=\linewidth]{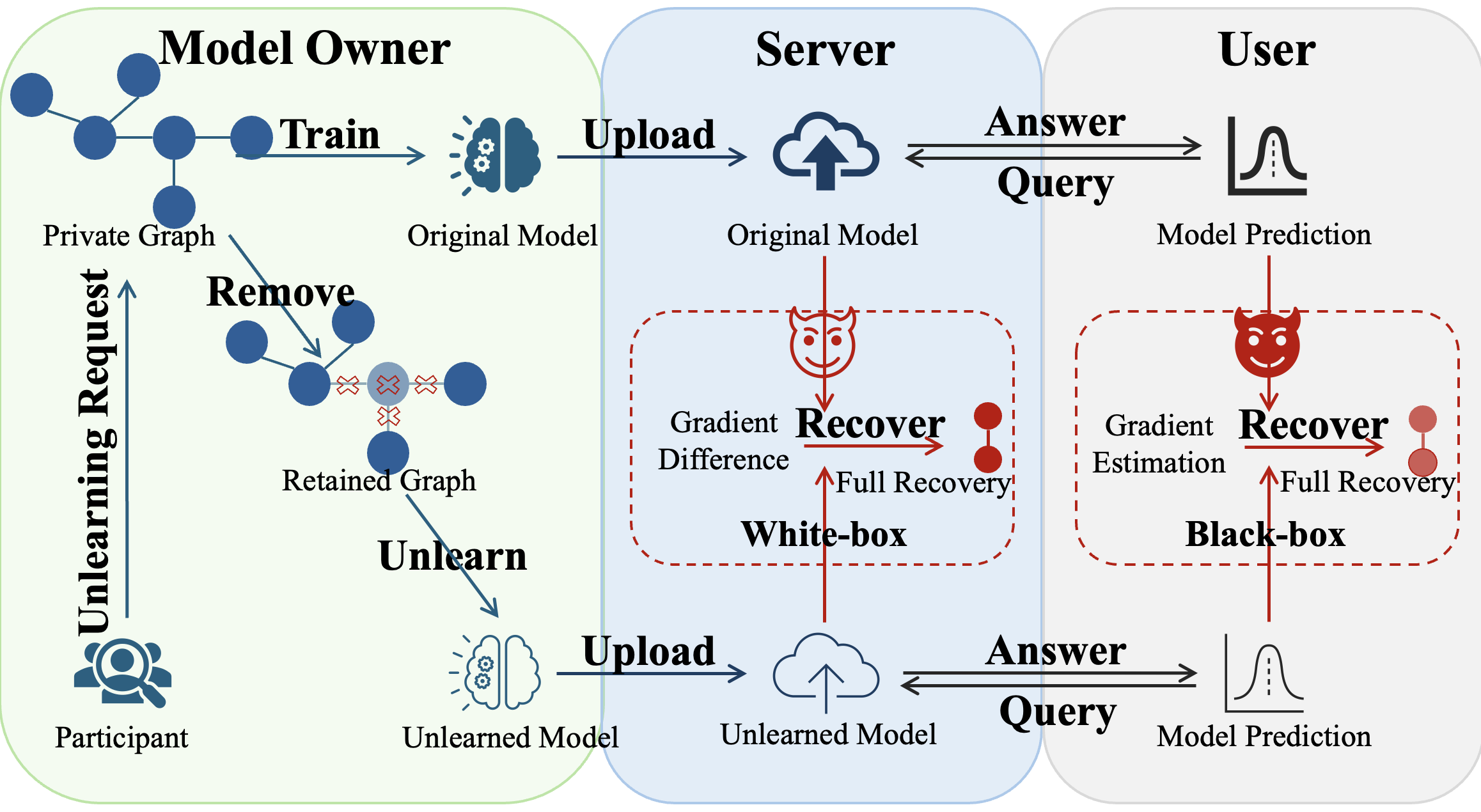}
  \captionsetup{justification=raggedright,singlelinecheck=false}
  \caption{An overview of GraphToxin. The white-box Gra- phToxin uses gradient difference for full unlearned graph recovery. The black-box GraphToxin first estimates gradie- nts and then exploits their difference for recovery.}
  \label{graphtoxin}
\end{figure}

We present the overview of GraphToxin in Figure \ref{graphtoxin}. GraphToxin contains three modules: \textit{Gradient Matching} to extract holistic information for the full unlearned graph recovery, \textit{Curvature Matching} to provide more fine-grained recovery signals and \textit{Feature Smoothness} to capture and enforce similarity among connected nodes within the unlearned graph. We detail each module as follows.

\subsubsection{\textbf{Gradient Matching}}
\hfill\break
\noindent\textbf{Design Rationale} Gradient matching is the foundation of gradient inversion attacks that are mainly tailored for federated learning. Given the gradients with respect to model parameters, gradient matching first initializes a dummy input and then iteratively optimizes it by minimizing the distance between the gradients induced by the dummy input and the observed model gradients. Since our attack settings are strictly data-free and only assume the adversary access to actual or estimated gradients of the original and unlearned models, gradient matching serves as a natural starting point.

However, gradient matching cannot be trivially applied to GU-FGRA. Though the original GNN contains the full information, without the unlearned model as a reference point, directly inverting the original model cannot isolate unlearned information from massive entangled global gradients. Even with identities of deleted nodes, attackers cannot decouple the precise structural and feature traces of the full learned graph from the massive background signals of the remaining graph. Moreover, due to the ambiguity of gradient-to-graph mappings, naively applying gradient matching to each gradient set are not guaranteed to align, hindering the reliable identification of unlearned graph components.



These limitations motivate a discrepancy-driven design that operates on gradient differences \cite{first_unlearn_attack} between the original and unlearned GNNs, rather than on separate gradient sets.

\noindent\textbf{Gradient Matching in GU-FGRA} For clarity, we begin with the single-node removal setting and discuss the extension to multiple-node removal in a separate section. 

Formally, let $\nabla\mathcal{L}$ and $\nabla\mathcal{L}^{-v}$ denote the gradients of the original and unlearned GNNs, respectively, where $-v$ represents an unlearning request on a single node $v$. We define the observed gradient difference as $\Delta\nabla\mathcal{L}=\nabla\mathcal{L}-\nabla\mathcal{L}^{-v}$, which captures unlearning-induced signals associated with the removed node $v$ and its affected neighborhoods $\mathcal{N}^1_{v}$. The adversary first initializes a dummy graph with the fixed unlearning graph size, and then optimizes it by minimizing the distance between the dummy and observed gradient difference to recover the full unlearned graph $\mathcal{G}_{u_{v}}=(\mathcal{V}_{u_{v}}, \mathcal{E}_{u_{v}}, \mathcal{X}_{u_{v}})$, where $\mathcal{V}_{u_{v}}=[v,\mathcal{N}^{1}_{v}]$. The attack objective is represented as:
\begin{equation}
    \tilde{\mathcal{G}}_{u_{v}} = \arg\min\limits_{\tilde{\mathcal{G}}_{u_{v}}}\mathcal{D}(\Delta\tilde{\nabla}\mathcal{L},\Delta\nabla\mathcal{L})
\end{equation}
where $\Delta\tilde{\nabla}\mathcal{L}$ denotes the gradients induced by the dummy graph.
For the choice of the distance function $\mathcal{D}$, existing gradient matching commonly uses cosine similarity \cite{first_gradient, first_unlearn_attack} or L2 distance \cite{deep_leakage_gradients}. Some prior studies favor L2 distance for its superior attack performance while others \cite{fastconvergentgradient} show that cosine similarity correlates more strongly with reconstruction quality. Inspired by Fan et al. \cite{realistic_gradient}, we combine both metrics to leverage their complementary strengths. Specifically, we adopt Mean Squared Error (MSE), a normalized form of L2 distance, to penalize the absolute deviations and mitigate the influence of magnitude amplification in the high-dimensional gradient space. Meanwhile, cosine similarity is used for direction consistency and semantic alignment. Following Fan et al. \cite{realistic_gradient}, we treat these two losses equally:
\begin{equation}
\begin{split}
    \mathfrak{L}_{grad} & = 1-\frac{\langle\Delta\tilde{\nabla}\mathcal{L},\Delta\nabla\mathcal{L}\rangle}{\|\Delta\tilde{\nabla}\mathcal{L}\|\cdot\|\Delta\nabla\mathcal{L}\|} + \frac{1}{M}\sum\limits_{i=1}^{M}\left((\Delta\tilde{\nabla}\mathcal{L})_i-(\Delta\nabla\mathcal{L})_i\right)^2
\end{split}
\end{equation}
where $M$ is the number of elements in the gradients.
\subsubsection{\textbf{Curvature Matching}} 
\hfill\break
\noindent\textbf{Motivation} Gradient matching provides holistic information for full unlearned graph recovery, however, it is not sufficient to extract more fine-grained recovery signals. The following illustrative example demonstrates how gradient matching leads to ambiguity and motivates the design of curvature matching.

\begin{figure}[htbp]
  \centering
  \includegraphics[width=\linewidth]{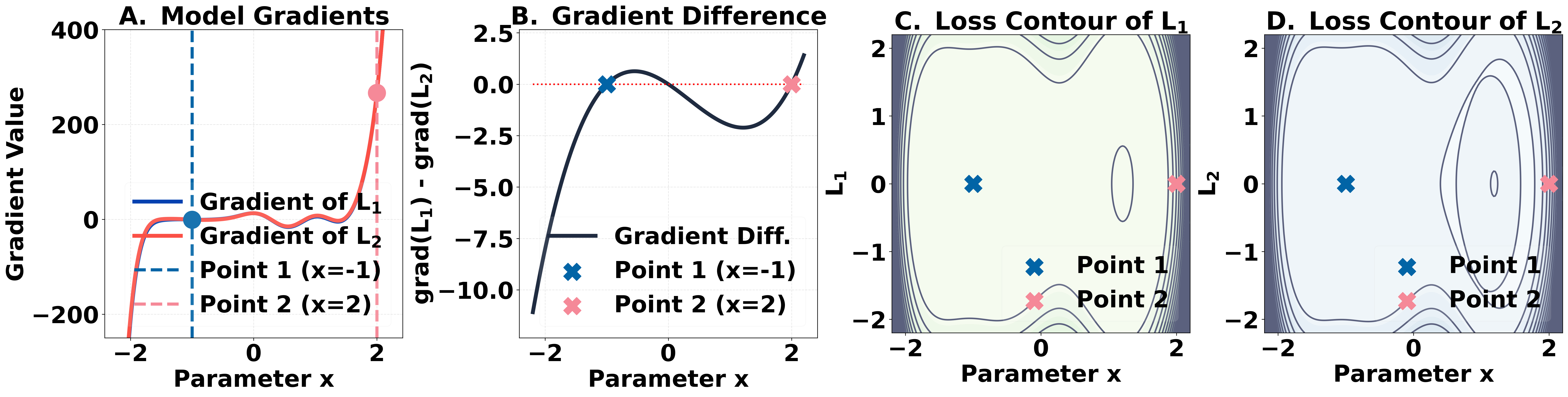}
  \captionsetup{justification=centering,singlelinecheck=false}
  \caption{Example of the Ambiguity of Gradient Matching} 
  \label{gradient_match_insufficient}
\end{figure}

As depicted in Figure \ref{gradient_match_insufficient}, we consider a convex loss function $\mathcal{L}_1$ and obtain $\mathcal{L}_2$ by adding noises and slightly changing the polynomial coefficients of $\mathcal{L}_1$ to simulate the effect of unlearning. As shown in Fig.\ref{gradient_match_insufficient}.A and B, the two losses induce nearly identical gradients and small gradient difference. However, we identify two distinct points (marked in blue and pink) that produce the identical gradient difference, yet exhibit the substantially different curvatures of the loss landscape as shown in Fig.\ref{gradient_match_insufficient}.C and D. This example highlights a fundamental limitation of gradient matching: \textit{multiple candidates can satisfy the same gradient difference constraint}. In graph learning, this ambiguity is further amplified by the permutation invariance of GNNs, under which different graph structures can yield identical information propagation and gradients. Moreover, GU-FGRA aims to recover not only the removed node and its incident edges, but also the features of affected neighboring nodes. While gradient difference can reflect the removal of a node and local connectivity changes, they impose insufficient constraints to uniquely characterize neighboring node features. To overcome this ambiguity and provide fine-grained signals for full unlearned graph recovery, we next introduce a novel curvature matching module. 


\noindent\textbf{Curvature Matching}
We treat curvature matching as a curvature-aware constraint operating on gradient difference. Detailed assumptions and proof are provided in Appendix \ref{assu_proof}. 

\begin{theorem}[\textbf{Curvature-aware difference Matching}]
\label{theorem_main}
    Under standard smoothness and local approximation assumptions, minimizing the gradient difference reduces to minimizing the curvature-weighted difference $(\Delta\tilde{\nabla}\mathcal{L}-\Delta\nabla\mathcal{L})^TH^{-1}(\Delta\tilde{\nabla}\mathcal{L}-\Delta\nabla\mathcal{L})$.
\end{theorem}

In practice, the Hessian matrix $H$ is computationally infeasible as it requires calculating and storing second-order derivatives. We leverage the Fisher Information Matrix $F$ as the surrogate. Specifically, $F$ is defined as the covariance matrix of the gradient of the log-likelihood function with respect to model parameters $\theta$, quantifying the sensitivity of the model's predicted probability distribution to small changes in $\theta$. According to Rame et al. \cite{fisher}, $F$ approximates $H$ with probably bounded errors under regular conditions. However, since the full $F$ involves an expectation over all possible outputs, the empirical Fisher Information $\tilde{F}$ is typically utilized as an efficient approximation \cite{empirical_fisher}. Formally, $\tilde{F}$ is denoted as:
\begin{equation}
\label{empirical}
    \tilde{F}=\frac{1}{N_u}\sum\limits_{i=1}^{N_u}(\nabla \log p(Y_i|\tilde{\mathcal{G}}_u,\theta))(\nabla \log p(Y_i|\tilde{\mathcal{G}}_u,\theta))^T
\end{equation}
where $N_u=|\mathcal{V}_{u}|$ is the node size of $\mathcal{G}_u$.

We then replace $H$ with $\tilde{F}$ in $(\Delta\tilde{\nabla}\mathcal{L}-\Delta\nabla\mathcal{L})^TH^{-1}(\Delta\tilde{\nabla}\mathcal{L}-\Delta\nabla\mathcal{L})$ and derive the following curvature matching loss:
\begin{equation}
    \mathfrak{L}_{curv}=(\Delta\tilde{\nabla}\mathcal{L}-\Delta\nabla\mathcal{L})^T\tilde{F}^{-1}(\Delta\tilde{\nabla}\mathcal{L}-\Delta\nabla\mathcal{L})
\end{equation}

$\mathfrak{L}_{curv}$ naturally encodes the curvature of the loss landscape, providing more fine-grained signals for full unlearned graph recovery. Not all gradient difference are equally informative: some may stem from noises or redundant parameterization, while others are highly indicative of the ground-truth features and underlying graph structures. By incorporating $\tilde{F}$, $\mathfrak{L}_{curv}$ downweights the directions of high curvature, thereby biasing the optimization toward smoother and more stable directions in parameter space. 
Furthermore, $\tilde{F}$ is defined with respect to model predictions and thus implicitly reflects how node features and graph structures are propagated through message passing and projected into posterior probabilities. This complementary perspective provides sensitivity-aware guidance that better aligns with the model’s local behaviors. 
\subsubsection{\textbf{Feature Smoothness}}
Most real-world graphs follow the homophily principle, where neighboring nodes tend to share similar features. Graph Dirichlet energy \cite{smooth_sparse, graph_mia, unlearning_inversion_attacks_graph} explicitly captures and quantifies this feature smoothness:
\begin{equation}
    \mathfrak{L}_{smooth}=tr(X^TLX)
\end{equation}
where $tr$ is the trace of a matrix, $L=D-A$ is the laplacian matrix and $D$ is the diagonal degree matrix.  

This regularization penalizes sharp feature differences across connected nodes and suppresses degenerate and noisy solutions for realistic full graph recovery. 

\subsubsection{\textbf{Loss Function}}
The final objective of GraphToxin is formalized as:
\begin{equation}
    \mathfrak{L}_{white}=\mathfrak{L}_{grad} + \alpha_1\mathfrak{L}_{curv} + \alpha_2\mathfrak{L}_{smooth}
\end{equation}
where $\alpha_1$ weights the curvature-aware loss while $\alpha_2$ controls feature smoothness regularization. We refer to them as the Fisher coefficient and the Laplacian coefficient, respectively.

\subsection{Extensions: Multiple-Node Removal and Black-box GraphToxin}
\label{black_box}

\subsubsection{Multiple-Node Removal}
For multiple-node removal, the topology of full unlearned graph $\mathcal{G}_u$ is significantly more complex due to the interactions of removed nodes and their affected neighbors.

To simultaneously reconstruct node features and such more complicated graph topology, we first initialize the topological structure by sampling a fixed number of edges using Bernoulli distribution \cite{gradient_attack_gnn}, marked as $\tilde{A}_0\sim Bernoulli(p)$, where $p=\frac{|\mathcal{E}_{u}|}{\binom{|\mathcal{V}_{u}|}{2}}$ controls the expected edge density and aligns with the known edge budget. Next, we adopt the same final objective $\mathfrak{L}_{white}$ as in the single-node removal case, while additionally updating the adjacency matrix $\tilde{A}$ via projected gradient descent \cite{gradient_attack_gnn}. Each entry $(i,j)$ of $\tilde{A}$ is updated through the entry-wise projection operator as:
\begin{equation}
\hat{A}_{ij}=\text{proj}_{[0,1]}(\tilde{A}_{ij})=\left\{
\begin{array}{ll}
    1, & \tilde{A}_{ij}\geq1 \\
    0, & \tilde{A}_{ij}\leq0 \\
    \tilde{A}_{ij}, & \text{otherwise}
\end{array} \right.
\end{equation}
We then repeat the sampling process using the learned edge probabilities from the last iteration to generate the final reconstructed adjacency matrix $\hat{A}$.

\subsubsection{Black-box GraphToxin}
Distinct from the white-box setting, black-box GraphToxin cannot access gradients and relies solely on posterior probabilities. Despite this constraint, the attack remains feasible in practice. The adversary first obtains surrogate models of the original and unlearned GNNs ($\mathcal{F}_o^{\prime}$ and $\mathcal{F}_u^{\prime}$) via an existing data-free black-box model extraction attack \cite{data_free_msa_gnn}. We
refer the interested readers to Zhuang et al. \cite{data_free_msa_gnn} for additional details.

Given copycats, GraphToxin exploits the gradient difference between these copycats using the same objective $\mathfrak{L}_{white}$. However, the reconstruction quality highly depends on the stealing quality. Imperfect extraction may introduce additional noise into the estimated gradients. We incorporate a semantic calibration loss to mitigate this effect:
\begin{equation}
\mathfrak{L}_{seman}=CE(\tilde{Y}_u, Y_u)=-\sum_{i=1}^{|\mathcal{V}_{u}|}Y_{u_i}\cdot\log p_{u_i}
\end{equation}
where $CE$ is the cross-entropy loss to minimize the discrepancies between the actual node labels and the model predictions of the dummy graph $\tilde{\mathcal{G}}_u$ from the copycat $\mathcal{F}_o^{\prime}$. 

The final objective for black-box GraphToxin is:
\begin{equation}
    \mathfrak{L}_{black}=\mathfrak{L}_{grad} + \alpha_1\mathfrak{L}_{curv} + \alpha_2\mathfrak{L}_{smooth}+\alpha_3\mathfrak{L}_{seman}
\end{equation}
where $\alpha_3$ controls the contribution of semantic calibration, referred as the calibration coefficient.

\subsection{Evaluation Framework Design}
\subsubsection{Worst-case Scenario}
Existing GU studies evaluate unlearning effectiveness under random node removals, but fail to examine worst-case scenarios where highly influential nodes are removed. This worst-case removal serves as a stress test rather than a favorable condition for the attack. Removing influential nodes indeed induces larger parameter changes, while simultaneously increases reconstruction difficulty. These nodes typically possess high degrees, amplifying multi-hop dependencies and thereby complicating the attribution of entangled features and topologies. Moreover, worst-case settings align with realistic attacker incentives. Nodes with high influence, such as celebrities in social networks, are more likely to attract stronger attack interests than normal users. 

For single-node removal, we unlearn the top 10\% nodes with the highest degrees from the training graph and report their average attack results to assess the worst-case scenarios, while in multiple-node removal, the top 25, 50, 75 nodes with the highest degrees are selected and partitioned into 5 non-overlapping pairs based on their degree ranking. 




\subsubsection{Feature-level Evaluation}
Feature-level evaluation measures the quality of features within the recovered unlearned graph $\tilde{\mathcal{G}}_u$. 

\noindent\textbf{Root Normalized Mean Squared Error (RNMSE)} We follow Sinha et al. \cite{gradient_attack_gnn} and leverage RNMSE to quantify the discrepancies between the recovered and actual node features. 
\begin{equation}
    \text{RNMSE}(\tilde{X}_v,X_v)=\frac{\| \tilde{X}_v-X_v\|}{\|X_v\|}
\end{equation}
We finally report the average RNMSE of all nodes in $\tilde{\mathcal{G}}_u$. While RNMSE is a standard metric for feature recovery, it does not fully capture structural similarity or functional consistency, as we empirically demonstrate in Section \ref{worst-analysis}.

\subsubsection{Global-level Evaluation}
Global-level evaluation assesses the preservation of joint feature–topology dependencies of $\tilde{\mathcal{G}}_u$. 

For single-node removal, since the graph topology is fixed, topology-level evaluation is unnecessary. For multiple-node removal, traditional link-wise reconstruction metrics like Area under the ROC Curve (AUROC) and Average Precision (AP) fail to capture holistic structural patterns. Therefore, we adopt the following global-level evaluation metrics:

\noindent\textbf{Embedding Distance (ED)} Node embeddings encode both node features and multi-hop structural patterns. Consequently, we measure the similarity between the embeddings of the recovered and ground-truth unlearned graphs, which are both extracted from the original GNN model, denoted as $\tilde{H}^k,H^k$, respectively. Here, $k$ is the number of GNN layers, usually $k=2$. We leverage the squared quadratic Wasserstein distance (p=2) \cite{vaserstein1969markov} to quantify this similarity, defined as $\mathrm{ED}=W_2^2(\tilde{H}^k,H^k)$.

\noindent\textbf{Pyramid Match Graph Kernel (PMGK)} PMGK is a robust measure for graph matching \cite{pmgk}, which captures structural similarity across multiple scales. Specifically, we first discretize continuous node embeddings of the recovered and ground-truth graphs into a set of discrete labels via K-means clustering, thereby grouping nodes with similar structural or semantic patterns. We then create a multi-resolution histogram of these labels of each graph and then take a weighted sum of the intersections of these two histograms. In our experiment, PMGK is computed using the grakel library \cite{grakel}.

\subsubsection{Performance-level Evaluation} Performance-level evaluation examines whether the recovered unlearned graph is functionally equivalent to the full unlearned graph when processed by the same GNN. This level of evaluation is essential as feature and structural similarities alone cannot guarantee the similar model behaviors.

\noindent\textbf{Attack Accuracy (ATT. ACC)} Attack accuracy measures whether the recovered graph leads to the same predicted labels as the ground-truth unlearned graph under the same GNN.
\begin{equation}
\mathrm{ATT.ACC}
=
\frac{1}{\left|\mathcal{V}_{u}\right|}
\sum_{v \in \mathcal{V}_{u}}
\mathbb{I}
\Big[
\arg\max \tilde{p}_v
=
\arg\max p_v
\Big]
\end{equation}
where $\mathbb{I}[\cdot]$ is the indicator function and $p$ is the posterior probability produced by  the original GNN $\mathcal{F}_o$.

\noindent\textbf{Attack Fidelity (ATT. FID)} Attack fidelity quantifies how accurately the recovered graph reproduces the model-specific predictions of the full ground-truth unlearned graph. 
\begin{equation}
\mathrm{ATT.FID}
=
\frac{1}{\left|\mathcal{V}_{u}\right|}
\sum_{v \in \mathcal{V}_{u}}
\mathbb{I}
\Big[
\arg\max \tilde{p}_v
=
\arg\max \hat{p}_v
\Big]
\end{equation}
where $\hat{p}_v$ denotes the posterior probability vectors produced by the original GNN on the ground-truth unlearned graph.

\noindent\textbf{Posterior Wasserstein Distance (PWD)} 
To further capture the confidence alignment at the distribution level, we measure the distance between posterior distributions induced by the recovered and ground-truth unlearned graphs using the squared quadratic Wasserstein distance as $\mathrm{PWD}
=
W_2^2(\tilde{p}, \hat{p})$.
\section{Experimental Validation}
\label{sec:exp}
In this section, we conduct empirical experiments across a range of GNN variants, graph unlearning methods and graph datasets to answer our three primary research questions. Specifically, we investigate the following key subquestions:

\noindent\textbf{SQ1}: Why is comprehensive evaluation necessary?

\noindent\textbf{SQ2}: How GraphToxin performs for single-node removal?

\noindent\textbf{SQ3}: Can GraphToxin flexibly adapt to different graph unlearning methods and GNN backbones?

\noindent\textbf{SQ4}: How GraphToxin performs in worst-case scenarios?

\noindent\textbf{SQ5}: Which impact factors dominate GraphToxin? 

\noindent\textbf{SQ6}: How GraphToxin performs in multiple-node removal?

\noindent\textbf{SQ7}: Can GraphToxin be extended to the black-box setting?

\noindent\textbf{SQ8}: How resilient is GraphToxin against defenses?

\subsection{Setups}

\subsubsection{Datasets} We conduct experiments on three benchmark graph datasets: Cora, PubMed \cite{cora_pub_data}, and Amazon-Photo (Photo) \cite{amazon_photo}. Dataset details are provided in Appendix \ref{data}. 

\subsubsection{Unlearning Methods} We consider the following graph unlearning methods. (1) Exact unlearning: We directly retrain the GNNs from scratch (Retraining). (2) Approximate unlearning: We consider two state-of-the-art graph unlearning methods with distinct frameworks: MEGU \cite{MEGU} and ETR \cite{ETR}. MEGU is a learning-based method that employs predictive and unlearning modules to mitigate unlearning effects, while ETR is a training-free method to erase the influence of deleted nodes and rectify the collateral damage to their neighbors. The original training and graph unlearning results across different GNN backbones and graphs are detailed in Table \ref{data_perf} in Appendix \ref{origin}, which confirm the effectiveness of all considered graph unlearning methods. 

\subsubsection{GNN Variants} We use three classical GNN backbones: GCN \cite{gcn}, GraphSAGE \cite{sage} and SGC \cite{sgc}.

\subsubsection{Baselines} To the best of our knowledge, GraphToxin is the first GU-FGRA, there are no established baseline methods for comparison. Accordingly, we consider the following alternative baselines: (1) Random Graph (Rand.): Random graphs are generated from Gaussian noise without parameter updates, proving a lower-bound reference for any non-trivial reconstruction. (2) Learning with A Few Epochs (FewE.): GraphToxin is trained for only the first 1\% of the total training epochs, simulating a weak adversary with severely limited computational budgets. This baseline tests whether the attack's success depends on extensive optimization or on the structural leakage exploited by GraphToxin. (3) Adapted Gradient Inversion (GLG-GU): We extend the gradient inversion attack \cite{deep_leakage_gradients, gradient_attack_gnn, grain} to the GU setting. Specifically, we adapt the state-of-the-art GLG \cite{gradient_attack_gnn}, which recovers node features and topology using a similarity-based reconstruction loss, a feature smoothness constraint, and a Frobenius norm topology regularizer.


\subsubsection{Implementation}
All experiments are implemented on $4$ Nvidia A$100$ GPUs. Please refer to Appendix \ref{implement} for implementation details.

\subsection{Attack Performance (SQ1 - SQ3)}

\subsubsection{Necessity of Comprehensive Evaluation (SQ1)} 
We first examine the necessity of worst-case evaluation. From Table \ref{toxin_perf}, we observe that averaging results under random-node removal can obscure meaningful differences in reconstruction difficulty, motivating the need for worst-case evaluation. PubMed exemplifies this problem. Due to its high sparsity and limited label space, random-node removal often produces relatively simple recovery targets, under which even randomly generated graphs may achieve strong performance-level scores. This effect can artificially inflate average performance and mask the weakness of Rand. on more complex unlearning instances. In contrast, under the worst-case settings, the performance of Rand. significantly decreases, while GraphToxin remains effective, providing a more faithful assessment under challenging unlearning scenarios.

We further motivate a comprehensive evaluation framework for GU-FGRA. Existing attacks commonly rely on RNMSE for assessing feature reconstruction quality \cite{gradient_attack_gnn}. While RNMSE is a standard and necessary metric, it is insufficient when used in isolation.
Table \ref{toxin_perf} shows that even a randomly generated graph may achieve a relatively low RNMSE while performing poorly on performance-level metrics. This discrepancy indicates that RNMSE alone can obscure qualitative differences between the recovered and ground-truth unlearned graphs, which are further amplified by the message-passing paradigm in GNNs. Hence, more sensitive metrics are required to discern subtle variations. Specifically, when RNMSE values are nearly indistinguishable, ED can reveal substantial divergence in the joint feature–structure space. Furthermore, since single-node removal typically induces only subtle parameter changes, reconstruction results across different unlearning methods yield distinct values yet appear nearly identical on the scale of $1e\!-\!2$. In contrast, PMGK can reflect fine-grained changes that are imperceptible to other metrics.  


\subsubsection{Effectiveness (SQ2)} 
As shown in Table \ref{toxin_perf} and \ref{toxin_perf_app}, GraphToxin and its variant FewE. consistently and substantially outperforms baselines across \textit{all} experimental settings on nearly every reconstruction quality and performance-level evaluation metrics. The superiority of GraphToxin generalizes across diverse graphs and graph unlearning methods. In addition, GraphToxin's overall advantage holds under both random and worst-case node removal scenarios, highlighting its robustness. On performance-level metrics, GraphToxin achieves up to $\mathbf{43.88\%}$ improvement over the state-of-the-art adapted GLG on Cora, and up to $\mathbf{35.26\%}$ on Photo. 
%


\begin{table*}[htbp]
\caption{\centering \textbf{The Performance of GraphToxin for Single-node Removal (Backbone: GCN)}. Unit: \textbf{1e-2}. Arrow indicates the direction of better performance and the \textcolor{blue}{blue} color denotes the superb \textbf{`best'} results (in \textbf{bold}).}
\label{toxin_perf}
\resizebox{\linewidth}{!}{
\begin{tabular}{c|c|c|cccccc|cccccc}
\toprule
\multirow{2}{*}{\textbf{Dataset}} & \multirow{2}{*}{\textbf{GU}} & \multirow{2}{*}{\textbf{Baseline}} & \multicolumn{2}{c}{\textbf{RNMSE($\downarrow$)}} & \multicolumn{2}{c}{\textbf{ED($\downarrow$)}} & \multicolumn{2}{c|}{\textbf{PMGK($\uparrow$)}} & \multicolumn{2}{c}{\textbf{ATT. ACC($\uparrow$)}} & \multicolumn{2}{c}{\textbf{ATT. FID($\uparrow$)}} & \multicolumn{2}{c}{\textbf{PWD($\downarrow$)}} \\ \cline{4-15} 
                                  &                              &                                    & Random           & Worst           & Random          & Worst         & Random           & Worst           & Random          & Worst          & Random          & Worst          & Random          & Worst         \\ \hline
\multirow{9}{*}{\textbf{Cora}}    & \multirow{3}{*}{Retrain}       & Rand.                              &6.59                  &3.61                 &78.76                 &72.45               &84.35                  &68.35                 &6.79                 &9.18                &6.31                 &7.48                &82.79                 &80.23               \\
                                  &                              & FewE                               &5.08                  &2.84                 &73.11                 &65.96               &\textbf{84.62}                  &\textbf{68.79}                 &57.50                 &39.29                &56.67                 &37.98                &40.47                 &56.30               \\

                                  &                              & GLG-GU  &\textbf{3.82} &1.98 &96.83 &71.57 &79.74 &66.38 &28.33 &45.44 &25.71 &40.73 &68.54 &53.66
   \\
                                  &                              & Toxin                              &4.11                  &\textbf{1.95}                 &\textbf{71.38}                 &\textbf{52.35}               &84.05                  &67.21                 &\textcolor{blue}{\textbf{72.21}}                 &\textcolor{blue}{\textbf{84.72}}                &\textcolor{blue}{\textbf{69.00}}                 &\textcolor{blue}{\textbf{77.97}}                &\textcolor{blue}{\textbf{29.66}}                 &\textcolor{blue}{\textbf{22.35}}               \\ \cline{2-15} 
                                  & \multirow{3}{*}{MEGU}        & Rand.                              &6.59                  &3.61                 &78.76                 &72.45               &84.19                  &\textbf{68.64}                 &6.79                 &9.18                &6.31                 &7.48                &82.79                 &80.23               \\
                                  &                              & FewE                               &5.08                  &2.84                 &73.16                 &66.02               &84.11                  &67.73                 &57.50                 &39.29                &56.67                 &37.98                &40.48                 &56.31               \\
                                  &                              & GLG-GU  &\textbf{3.82} &\textbf{1.98} &96.83 &71.57 &80.32 &67.57 &28.33 &45.44 &25.71 &40.73 &68.54 &53.66   \\
                                  &                              & Toxin                              &4.23                  &2.10                 &\textbf{71.84}                 &\textbf{51.18}               &\textbf{84.30}                  &67.21                 &\textcolor{blue}{\textbf{71.14}}                 &\textcolor{blue}{\textbf{82.22}}                &\textcolor{blue}{\textbf{70.31}}                 &\textcolor{blue}{\textbf{77.85}}                &\textcolor{blue}{\textbf{30.22}}                 &\textcolor{blue}{\textbf{20.72}}               \\ \cline{2-15} 
                                  & \multirow{3}{*}{ETR}         & Rand.                              &6.59                  &3.61                 &78.76                 &72.43               &84.16                  &67.37                 &6.79                 &9.18                &6.31                 &7.48                &82.79                 &80.23               \\
                                  &                              & FewE                               &5.08                  &2.84                 &73.16                 &66.02               &\textbf{84.97}                  &\textbf{67.73}                 &57.50                 &39.29                &56.67                 &37.98                &40.48                 &56.31               \\
                                  &                              & GLG-GU  &\textbf{3.82} &\textbf{1.98} &96.83 &71.57 &80.09 &66.85 &28.33 &45.44 &25.71 &40.73 &68.54 &53.66   \\
                                  &                              & Toxin                              &4.23                  & 2.10                &\textbf{71.84}                 & \textbf{51.18}              &84.14                  &66.92                 &\textcolor{blue}{\textbf{71.14}}                 &\textcolor{blue}{\textbf{82.22}}                &\textcolor{blue}{\textbf{70.31}}                 &\textcolor{blue}{\textbf{77.85}}                &\textcolor{blue}{\textbf{30.22}}                 &\textcolor{blue}{\textbf{20.72}}               \\ \hline
\multirow{9}{*}{\textbf{Photo}}            & \multirow{3}{*}{Retrain}       & Rand.                              &\textcolor{blue}{\textbf{0.61}}                  &\textcolor{blue}{\textbf{0.31}}                 &92.58                 &107.83               &\textbf{69.33}                  &\textbf{63.29}                 &26.01                 &2.98                &25.20                 &0.93                &74.48                 &92.92               \\
                                  &                              & FewE                               &\textcolor{blue}{\textbf{0.61}}                  &\textcolor{blue}{\textbf{0.31}}                 &51.46                 & 68.21              &68.03                  &58.10                 & 67.44                &28.76                &68.02                 &28.74                &28.12                 &62.26               \\
                                  &                              & GLG-GU  &0.70 &0.35 &66.79 &66.96 &67.20 &53.40 &51.03 &58.09 &50.39 &56.04 &47.93 &48.42   \\
                                  &                              & Toxin                              &0.64                  &0.34                 &\textcolor{blue}{\textbf{44.49}}                 &\textcolor{blue}{\textbf{52.11}}               &66.33                  &49.65                 &\textcolor{blue}{\textbf{86.11}}                 &\textcolor{blue}{\textbf{85.95}}                &\textcolor{blue}{\textbf{85.65}}                 &\textcolor{blue}{\textbf{80.85}}                &\textcolor{blue}{\textbf{13.91}}                 &\textcolor{blue}{\textbf{30.36}}               \\ \cline{2-15} 
                                  & \multirow{3}{*}{MEGU}        & Rand.                              &\textcolor{blue}{\textbf{0.61}}                  &\textcolor{blue}{\textbf{0.31}}                 &92.58                 &107.83               &\textbf{68.42}                  &\textbf{63.22}                 &26.01                 &2.98                &25.20                 &0.93                &74.48                 &92.92               \\
                                  &                              & FewE                               &\textcolor{blue}{\textbf{0.61}}                  &\textcolor{blue}{\textbf{0.31}}                 &51.46                 &68.21               &68.08                 &57.81                 &67.44                 &28.76                &68.02                 &28.74                &28.12                 &62.26               \\
                                  &                              & GLG-GU  &0.70 &0.35 &66.79 &66.96 &66.86 &53.51 &51.03 &58.09 &50.39 &56.04 &47.93 &48.42   \\
                                  &                              & Toxin                              &0.64                  &0.34                 &\textcolor{blue}{\textbf{44.49}}                 &\textcolor{blue}{\textbf{51.83}}               &66.36                  &49.06                 &\textcolor{blue}{\textbf{86.11}}                 &\textcolor{blue}{\textbf{86.27}}                &\textcolor{blue}{\textbf{85.65}}                 &\textcolor{blue}{\textbf{80.66}}                &\textcolor{blue}{\textbf{13.91}}                 &\textcolor{blue}{\textbf{30.43}}               \\ \cline{2-15} 
                                  & \multirow{3}{*}{ETR}         & Rand.                              &\textcolor{blue}{\textbf{0.61}}                  &\textcolor{blue}{\textbf{0.31}}                 &92.58                 &107.83               &\textbf{68.84}                  &\textbf{63.48}                 &26.01                 &2.98                &25.20                 &0.93                &74.48                 &92.92               \\
                                  &                              & FewE                               &\textcolor{blue}{\textbf{0.61}}                  &\textcolor{blue}{\textbf{0.31}}                 &51.46                 &68.21               &68.60                  &58.02                 &67.44                 &28.76                &68.02                 &28.74                &28.12                 &62.26               \\
                                  &                              & GLG-GU  &0.70 &0.35 &66.79 &66.96 &67.65 &53.59 &51.03 &58.09 &50.39 &56.04 &47.93 &48.42   \\
                                  &                              & Toxin                              &\textcolor{blue}{\textbf{0.61}}                  &\textcolor{blue}{\textbf{0.31}}                 &\textcolor{blue}{\textbf{44.49}}                 &\textcolor{blue}{\textbf{51.83}}               &66.36                  &49.06                 &\textcolor{blue}{\textbf{86.11}}                 &\textcolor{blue}{\textbf{86.27}}                &\textcolor{blue}{\textbf{85.65}}                 &\textcolor{blue}{\textbf{80.66}}                &\textcolor{blue}{\textbf{13.91}}                 &\textcolor{blue}{\textbf{30.43}}               \\ \bottomrule
\end{tabular}}
\end{table*}

\subsubsection{Flexibility (SQ3)} We evaluate the flexibility of GraphToxin across different graph unlearning methods and GNN backbones.

\noindent\textbf{Graph Unlearning} The results in Table \ref{toxin_perf} show the stable and consistent performance of GraphToxin across \textit{all GU methods, even Retraining, a silver bullet for full removal}, we leave the explanations in Section \ref{ripple}. While minor variations are observed, these differences arise from the interaction between how unlearning signals are manifested in the parameter space and structural properties of the underlying graphs. Specifically, the attack against Retraining tends to induce larger parameter changes, which are particularly exploitable in small and sparse graphs such as Cora. In contrast, MEGU introduces learning-based modules to suppress large parameter deviations. This design mitigates the propagation of unlearning signals and can lead to reduced distinguishability under structure-sensitive metrics such as PMGK. ETR adopts a training-free and highly localized strategy that erases the influence of deleted nodes while rectifying collateral damage to neighboring nodes. Therefore, ETR can better preserve local structures, which may account for its stronger global-level alignment across diverse graphs.

\noindent\textbf{GNN Backbones} We next fix the unlearning target as Retraining and investigate the impact of different GNN backbones. Compared to the nearly identical performance observed across unlearning methods in Table \ref{toxin_perf}, \textit{GraphToxin exhibits distinct performance across GNN backbones, while still maintaining its overall superiority and generalization across all graph datasets} (Table \ref{toxin_perf_backbone}). Detailed results and analysis are provided in Appendix \ref{att_backbone} due to space constraints.



\subsection{Parameter Studies (SQ4 and SQ5)}

\subsubsection{Worst-case Analysis} 
\label{worst-analysis}
Previous subsections motivate the necessity of worst-case evaluation. We now analyze its impact on attack performance. As summarized in Table \ref{toxin_perf} and Table \ref{toxin_perf_backbone}, worst-case node removal leads to markedly different behaviors across graphs. On Cora, GraphToxin consistently achieves stronger performance under worst-case removal across nearly all metrics, except PMGK. In contrast, on PubMed and Photo, GraphToxin performs better under random-node removal for global and performance-level metrics. Similar trends are observed across different GNN backbones.  

\textit{This divergence highlights a trade-off in worst-case evaluation.} Removing high-degree nodes increases reconstruction difficulty by amplifying structural ambiguity and semantic misalignment. While moderate neighborhood expansion can yield more informative signals for recovery, these signals may be distorted by over-smoothing or losing crucial semantic and structural details during information propagation, particularly within heterogeneous neighborhoods.
\begin{figure*}[htbp!]
    \centering
    \subfloat{\includegraphics[width=0.5\linewidth]{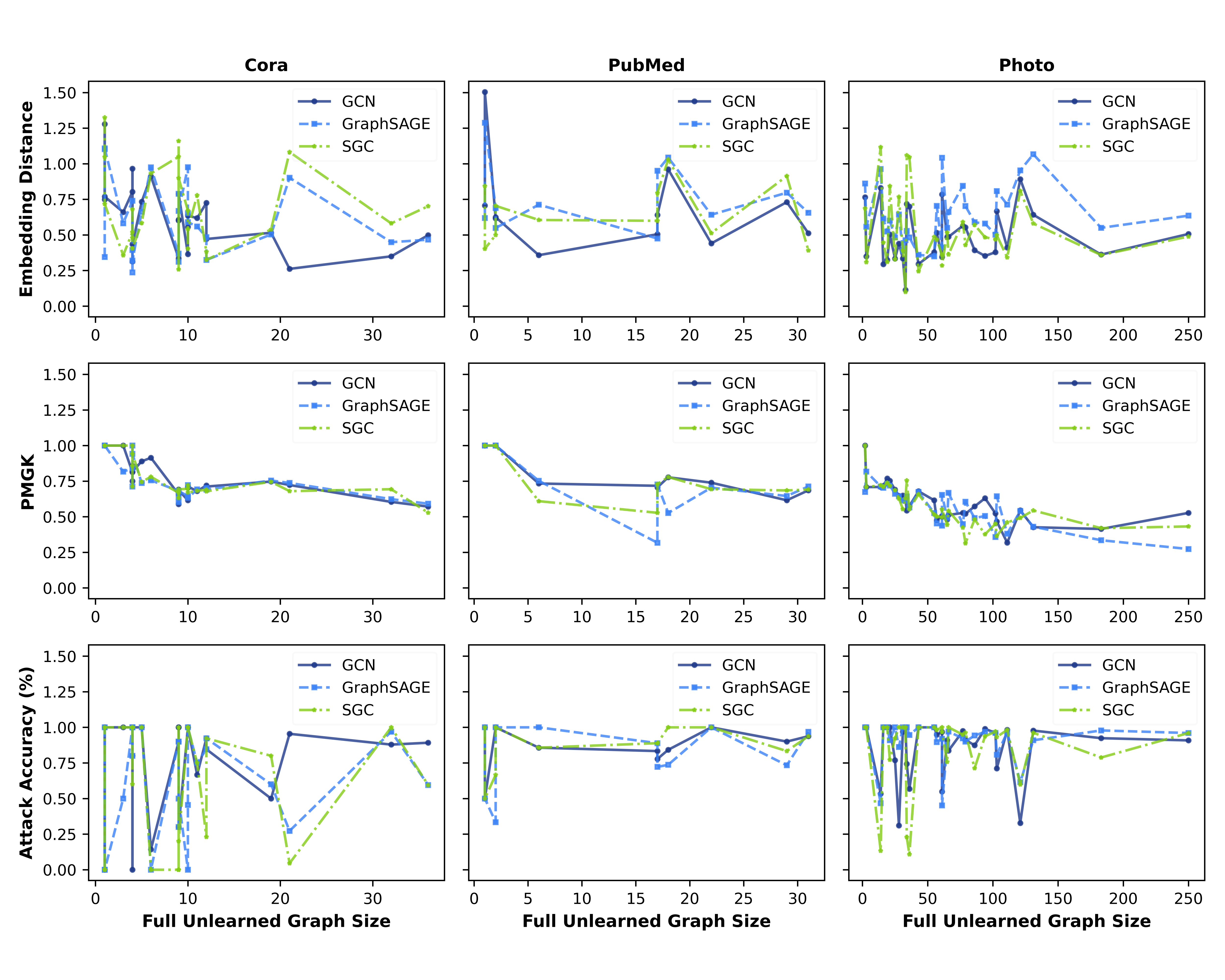}}
    \subfloat{\includegraphics[width=0.5\linewidth]{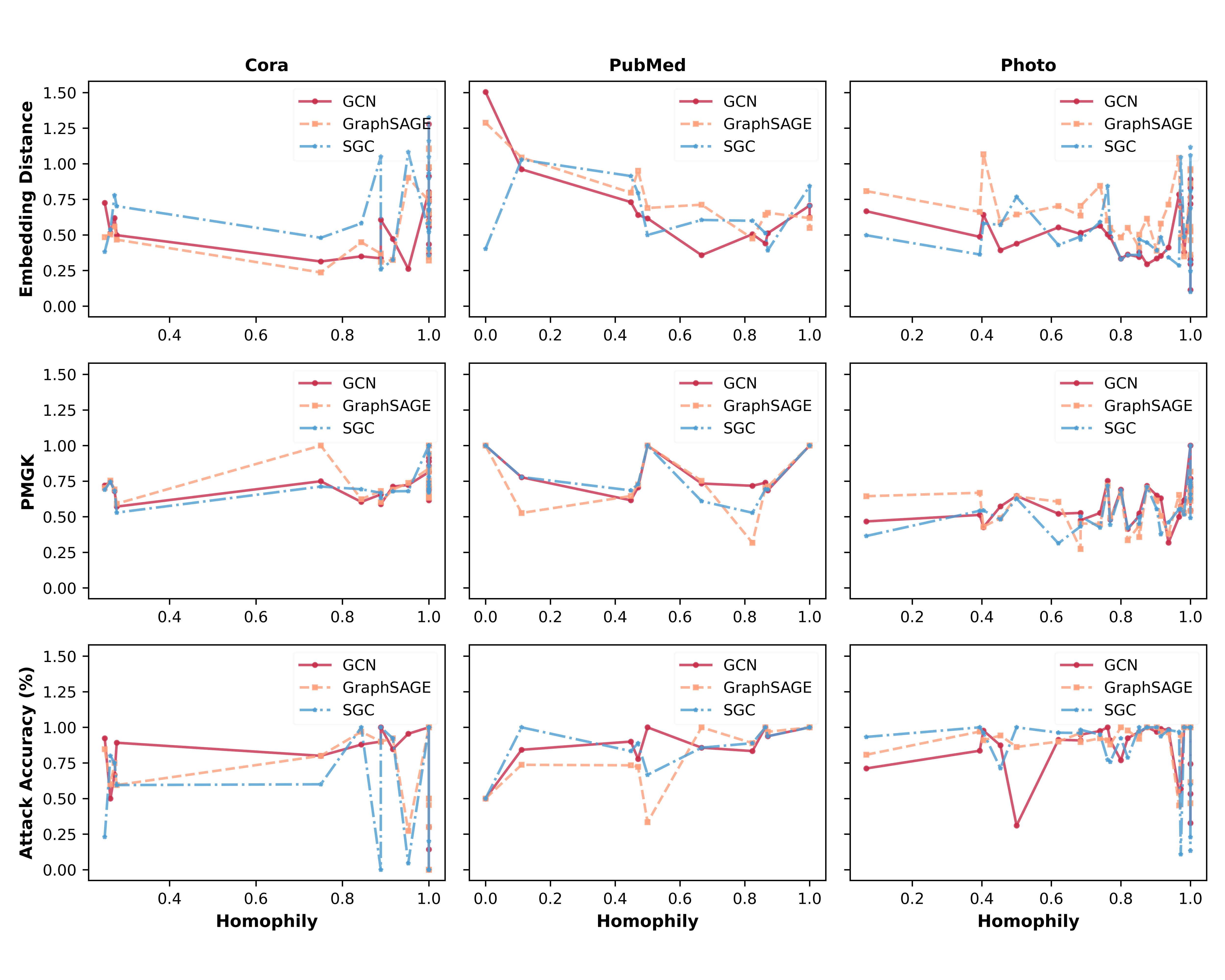}}
    \caption{Impact of the Full Unlearned Graph Size and Node Homophily}
    \label{node_size_homo}
\end{figure*}

\subsubsection{Unlearned Graph Size} 
Figure \ref{node_size_homo} shows that \textit{the attack performance is unstable when the unlearned graph is small. As the graph size increases, the performance stabilizes and gradually converges across different GNN backbones.} We provide the visualization of rest metrics and corresponding analysis in Appendix \ref{imp_graph_size}.


\subsubsection{Node Homophily} 
From Figure \ref{node_size_homo}, \textit{GraphToxin exhibits stable performance under varying homophily levels}, with GCN achieving the strongest performance across most metrics. As homophily approaches one, GraphSAGE and SGC diverge from GCN mainly in performance-level evaluation. Outside this range, the attack performance tends to converge. One possible explanation is that near-perfect homophilic neighborhoods produce indistinguishable node embeddings, thus leading to noninformative gradients for recovery. Similarly, the visualization and analysis of rest metrics are presented in Appendix \ref{imp_node_homo}. 


\subsubsection{Fisher Coefficient} 
\label{abs}
\begin{figure}[htbp]
    \centering
    \includegraphics[width=0.99\linewidth]{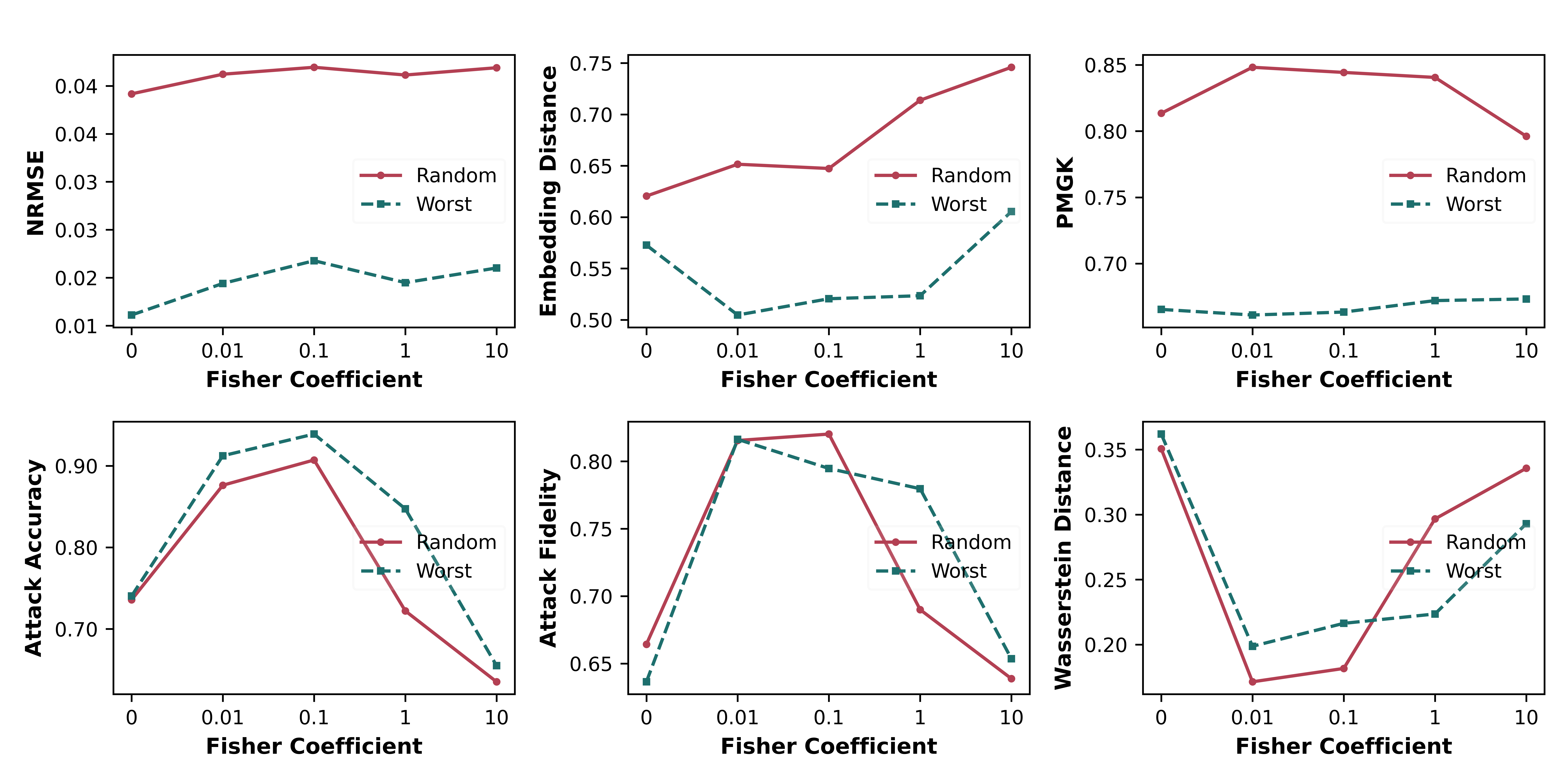}
    \caption{Impact of Fisher Coefficient}
    \label{fisher}
\end{figure}

We evaluate the effectiveness of curvature matching by varying its Fisher coefficient ($\alpha_1$) from $\{0, 0.01, 0.1, 1, 10\}$, where $\alpha_1 = 0$ serves as the ablation. Figure \ref{fisher} \textit{empirically confirms the effectiveness of curvature matching}, as the attack performance generally improves when the coefficient is non-zero. Consistent with previous findings, the results under worst-case removal typically exceed those under random removal. 
Additionally, the random case achieves higher attack fidelity when $\alpha_1 \in \{0.01, 0.1\}$ and lower posterior Wasserstein distance when $\alpha_1 \in \{0, 0.01, 0.1\}$. 
Overall, GraphToxin attains competitive performance when $\alpha_1\in \{0.01,0.1, 1\}$ and yields better results for both random and worst cases when $\alpha_1=0.1$.


\begin{figure}[htbp]
    \centering
    \includegraphics[width=0.99\linewidth]{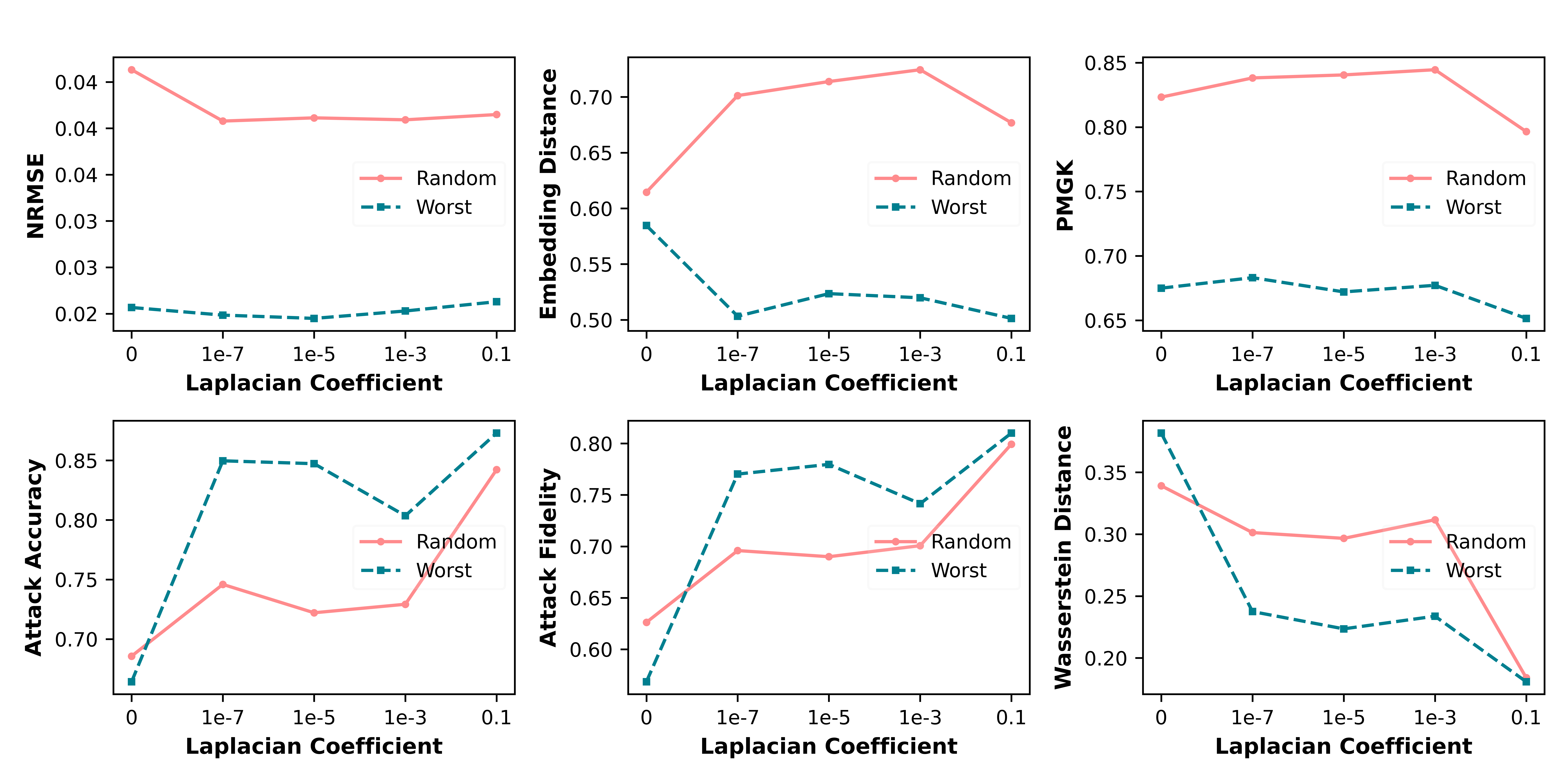}
    \caption{Impact of Laplacian Coefficient}
    \label{laplacian}
\end{figure}

\subsubsection{Laplacian Coefficient}
Figure \ref{laplacian} \textit{demonstrates the effectiveness of the feature smoothing module}. 
Similar to prior studies that set Laplacian coefficients ($\alpha_2$) to fairly small values \cite{gradient_attack_gnn, graph_mia}, we tune $\alpha_2\in\{0, 1e\!-\!7, 1e\!-\!5, 1e\!-\!3, 0.1\}$. We observe that the attack performance under the worst-case removal exceeds that under the random removal in nearly all metrics while the random case achieves stronger performance mainly in global and performance-level metrics. Overall, GraphToxin achieves superior performance when $\alpha_2\in\{1e\!-\!7, 1e\!-\!5\}$. 

\subsubsection{Label Noise}
\label{label_noise_sec}
\begin{table}[htbp]
\centering 
\caption{\centering \textbf{The Performance of GraphToxin under Label Noise (Dataset: Cora; Backbone: GCN; GU: Retraining)}} 
\label{toxin_perf_defense}
\resizebox{\linewidth}{!}{
\begin{tabular}{c|cccccc|cccccc}
\hline
\multirow{2}{*}{\textbf{Noise}} & \multicolumn{2}{c}{\textbf{RNMSE($\downarrow$)}} & \multicolumn{2}{c}{\textbf{ED($\downarrow$)}} & \multicolumn{2}{c|}{\textbf{PMGK($\uparrow$)}} & \multicolumn{2}{c}{\textbf{ACC($\uparrow$)}} & \multicolumn{2}{c}{\textbf{FID($\uparrow$)}} & \multicolumn{2}{c}{\textbf{PWD($\downarrow$)}} \\ \cline{2-13} 
                                                                  & Random           & Worst           & Random          & Worst         & Random           & Worst           & Random          & Worst          & Random          & Worst          & Random          & Worst         \\ \hline
$0\%$                           &4.11                  &1.95                 &71.38                 &52.35               &84.05                  &67.21                 &72.21                 &84.72                &69.00                &77.97                &29.66                 &22.35         \\ 

                                    $5\%$                           & 4.26             & 2.21            & 71.51           & 51.54         & 84.27            & \textbf{68.00}           & 72.93           & 81.15          & 70.31           & 77.09          & 30.07           & 21.09         \\ 

                                    $10\%$                           & 4.61             & 2.01            & 65.22           & \textbf{44.09}         & \textbf{84.52}            & 65.57           & 77.12           & \textbf{85.76}          & 73.18           & \textbf{78.37}          & 25.63           & \textbf{19.80}         \\ 

                                    $15\%$                           & 4.02             & 1.86            & 60.57           & 44.81         & 81.88            & 66.14           & 76.36           & 82.73          & 72.91           & 74.70          & 25.87           & 23.40         \\ 

                                    $20\%$                           & \textbf{3.23}             & \textbf{1.65}            & \textbf{55.43}           & 45.88         & 81.28            & 66.57           & \textbf{81.43}           & 76.13          & \textbf{76.19}           & 69.11          & \textbf{20.10}           & 26.04         \\ \hline
\end{tabular}}
\label{label_noise_table}
\end{table}

Existing label inference attacks can near-perfectly recover labels from model gradients \cite{deep_leakage_gradients, first_unlearn_attack}, therefore, we restrict our investigation under zero-to-modest levels of label noise. Given a noise ratio $\eta \in [0\%, 20\%]$, we randomly select a proportion $\eta$ of nodes and flip their ground-truth labels to other candidate classes. As shown in Table \ref{label_noise_table}, \textit{GraphToxin exhibits strong resilience against label noise}, particularly under random-node removal where performance improves consistently across most metrics as $\eta$ increases. In contrast, under the worst-case scenario, performance peaks around $\eta=10\%$. We attribute this counterintuitive performance gain to such noise acting as an implicit regularizer that smooths the loss surface and mitigates potential overfitting.

\subsection{Extensions (SQ6 and SQ7)} 
Due to space limitations, we provide detailed performance and analysis of the extensions to multiple-node removal and the black-box setting in Appendix \ref{multi_node} and \ref{black_box_toxin}, respectively.

\subsection{Discussion}

\subsubsection{Ripple Effects}
\label{ripple}
We now investigate why Retraining, widely regarded as the gold standard for full removal without any detectable traces, remains susceptible to GraphToxin. We posit that due to the ``ripple effects'' propagated through GNN message passing, even removing a single node induces structural footprints that persist in the model's behaviors. We adopt the posterior change $\Delta_p$ before and after unlearning as the footprint indicator for two reasons. First, any perturbation in the graph propagates to and manifests in posteriors. In contrast, raw gradients often exhibit large magnitudes that can mask the subtle perturbations induced by a single node deletion. Second, under the black-box setting, posteriors are the only signals accessible to an adversary, making them the most realistic surface for detecting exploitable leakage. We measure $\Delta_p$ using L2 distance. To further examine the propagation of these footprints, we also compute the mean L2 distance between layer-wise embeddings of the original and unlearned models.

\begin{figure}[htbp]
    \centering
    \includegraphics[width=0.99\linewidth]{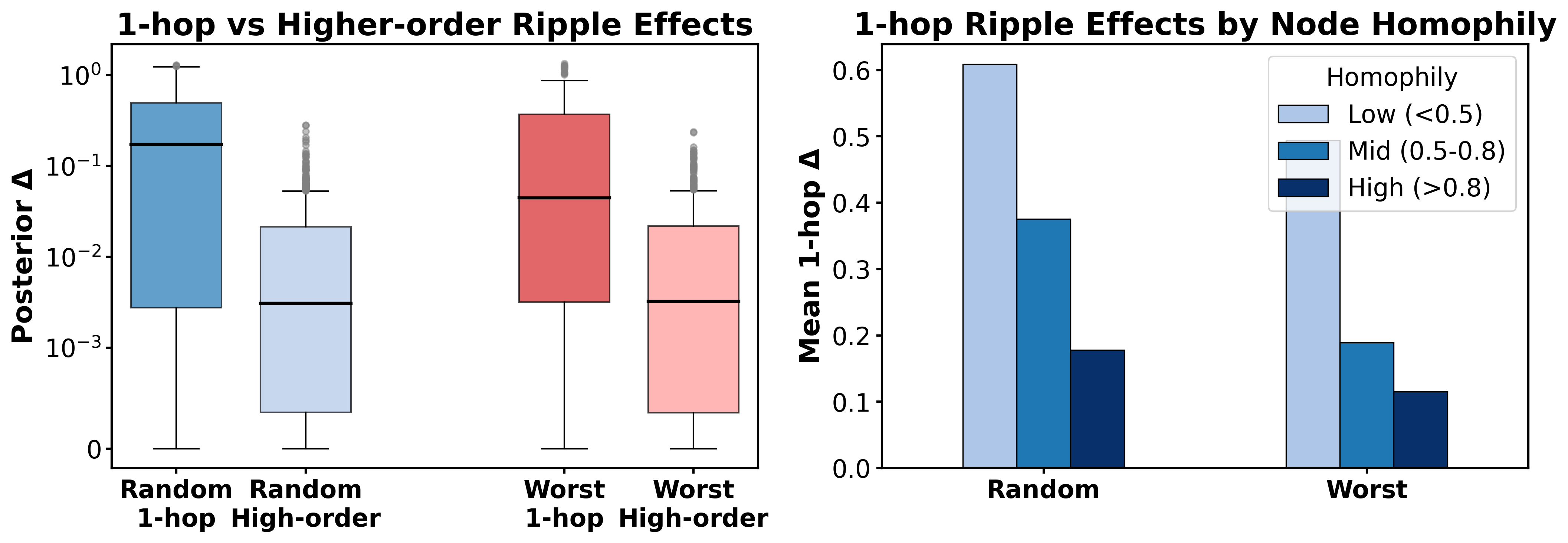}
    \caption{1-hop and Higher-order Ripple Effects}
    \label{one_hop_ripple}
\end{figure}

\begin{figure}[htbp]
    \centering
    \includegraphics[width=0.99\linewidth]{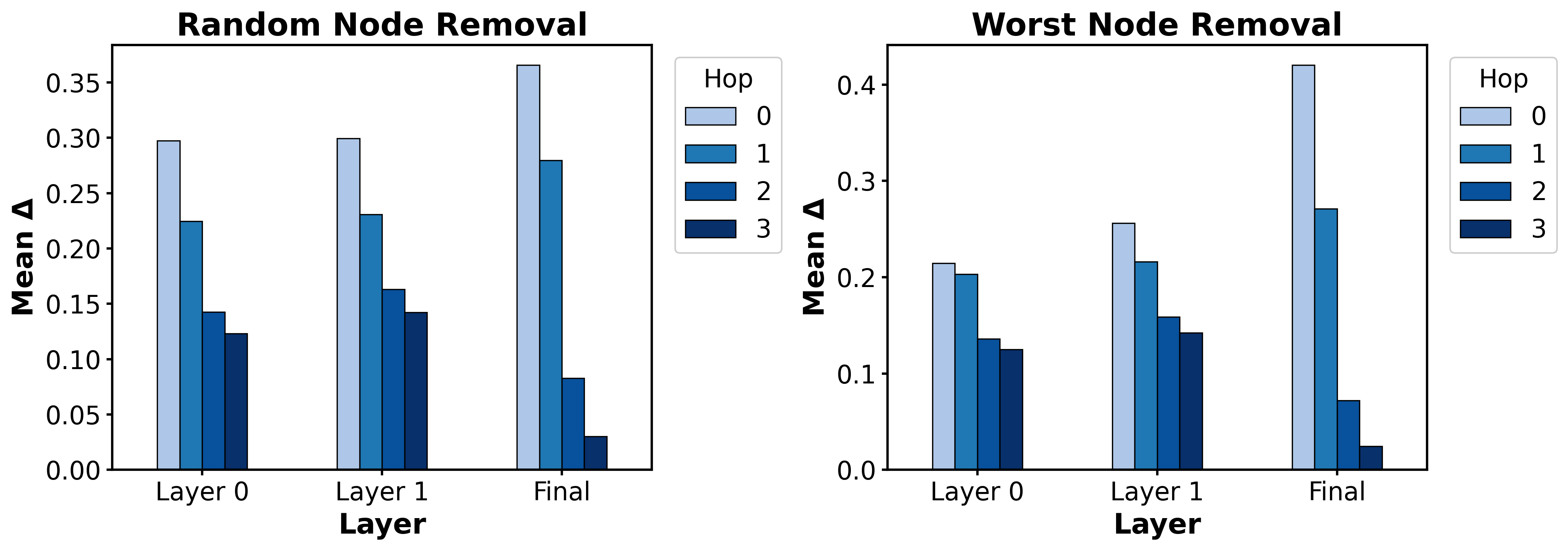}
    \caption{Impact of Layers in Ripple Effects}
    \label{multiple_hop_ripple}
\end{figure}

\begin{figure}[t]
\centering
\begin{minipage}[t]{0.48\linewidth}
    \centering
    \includegraphics[width=\linewidth]{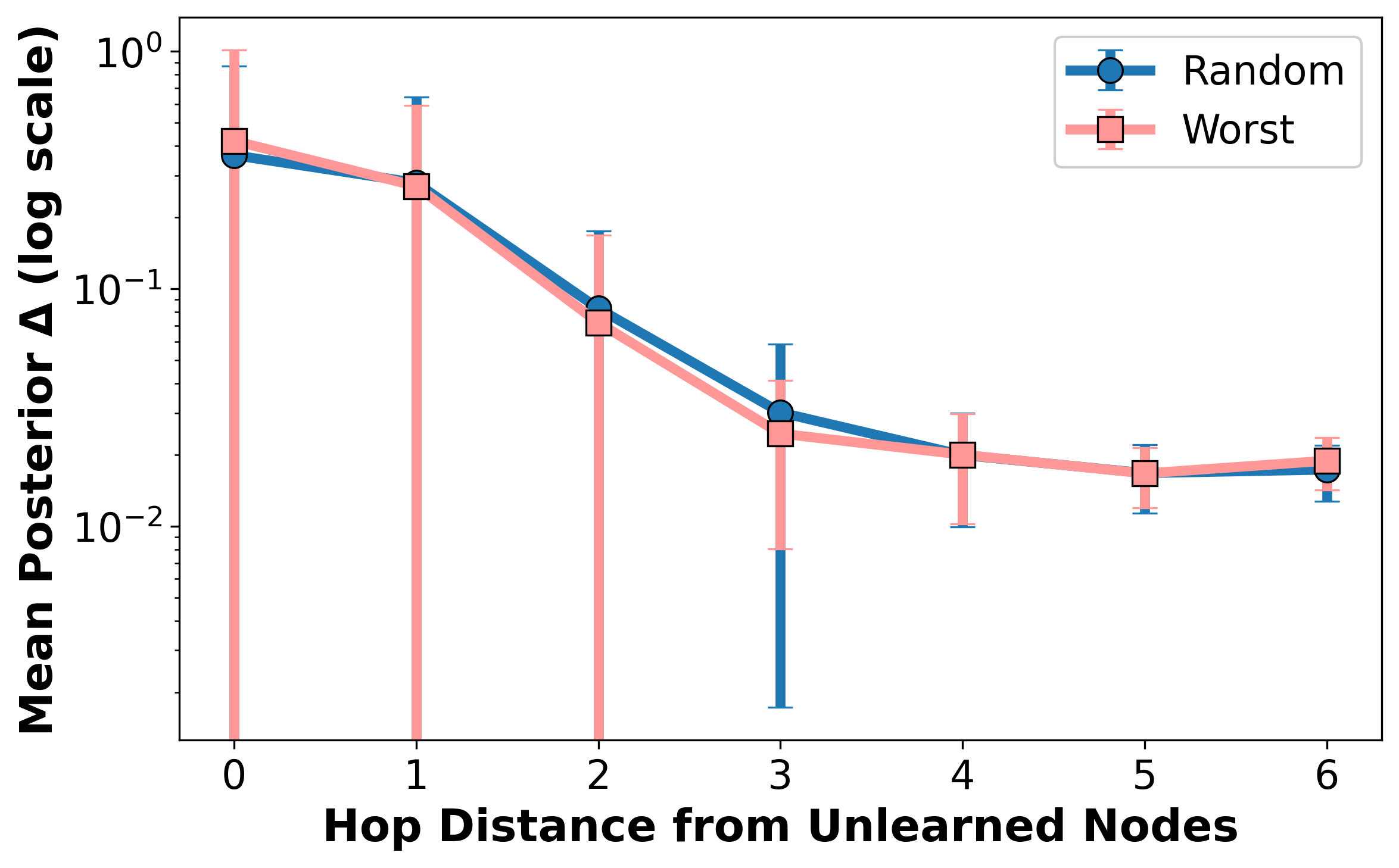}
    \caption{Multi-hop Ripple Effects}
    \label{multi_hop}
\end{minipage}
\hfill
\begin{minipage}[t]{0.48\linewidth}
    \centering
    \includegraphics[width=\linewidth]{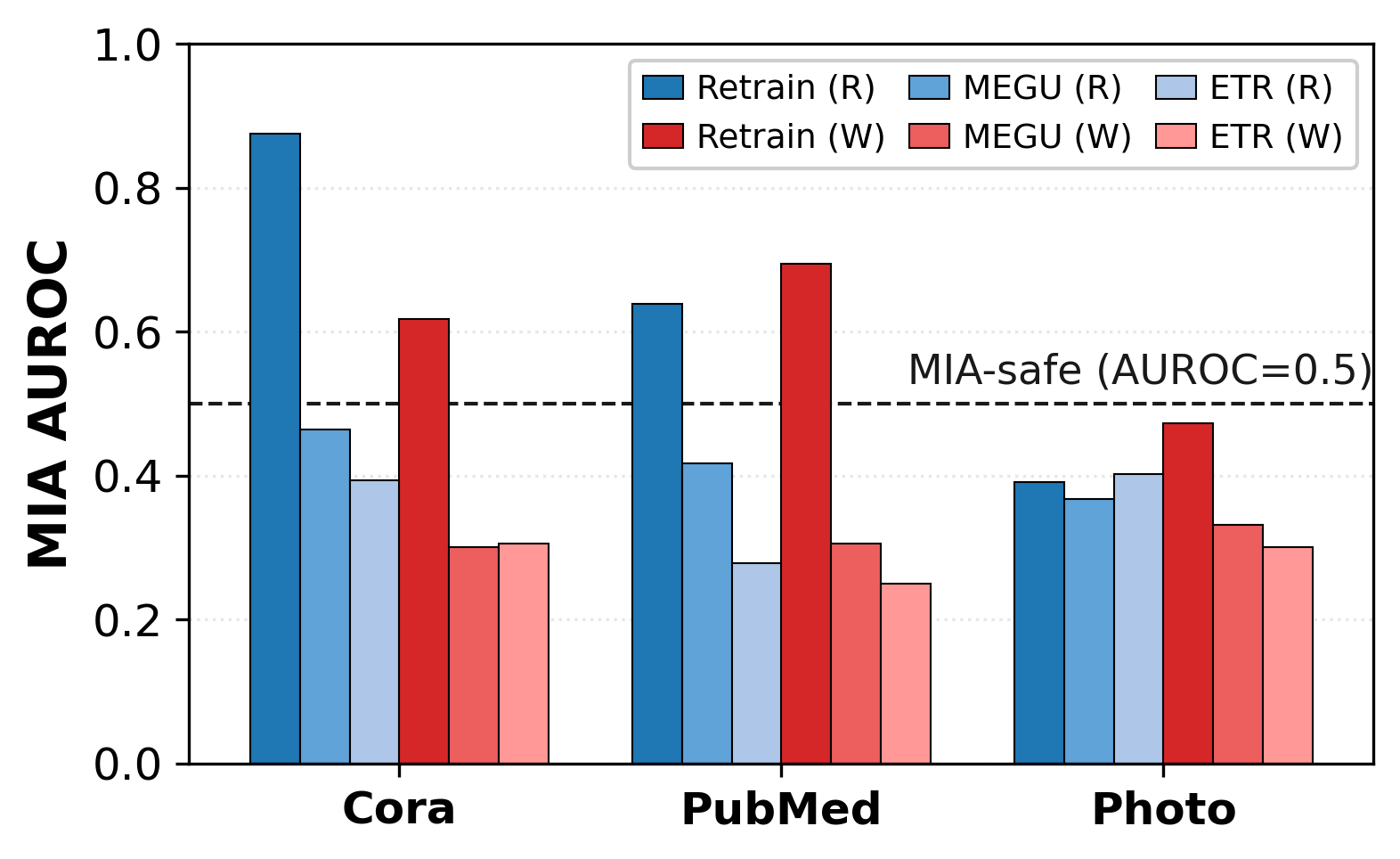}
    \caption{MIA AUROC across GU. AUROC near 0.5 indicates MIA-safe.}
    \label{fig:mia_auroc}
\end{minipage}
\label{fig:mia_combined}
\end{figure}

As shown in Figure \ref{one_hop_ripple}, structural footprints on deleted samples' 1-hop neighbors are more pronounced than that on higher-order neighbors, providing direct empirical evidence of ``ripple effects''. The footprint magnitude is inversely correlated with node homophily: deleted nodes in heterophilic regions trigger more significant changes than those in homophilic regions. From Figure \ref{multiple_hop_ripple}, we observe that the final output layer consistently amplifies the divergence for low-hop neighbors. Compared to the worst-case scenario, the random removal consistently yields larger 1-hop footprints than worst-case removal, a gap that persists across homophily levels and network layers. Figure \ref{multi_hop} demonstrates the ``ripple effect'' diminishes rapidly as the distance from the unlearned nodes increases, with random and worst-case removals exhibiting nearly identical decay curves. In summary, these observations empirically validate the existence of ``ripple effects'' in GU and characterize their structural propagation patterns, explaining why even Retraining fails to fully erase information from the graph. Due to space limitations, we put the results and analysis of deletion size in Appendix \ref{ripple_size}.

\subsubsection{Membership Inference Attack (MIA)}
The GU benchmark \cite{gu_bench} has validated that MEGU and ETR are MIA-safe under random-node removal. Following the default MIA setting on GU \cite{mia_gnn_2021,gu_bench}, we implement a shadow model-based MIA that exploits the discrepancy in posterior changes between member and non-member nodes, and report AUROC as the MIA evaluation metric. Intuitively, member nodes incur larger posterior changes than non-member nodes. When AUROC approaches $0.5$, the membership inference is indistinguishable from random guessing. 

From Figure \ref{fig:mia_auroc}, Retraining consistently exhibits higher MIA AUROC, suggesting that exact unlearning inadvertently creates behavioral signatures that an adversary can exploit. The susceptibility to MIA under random and worst-case removals varies across settings, with neither showing universal dominance. Critically, approximate GU tend to satisfy MIA-based verification especially on Cora and Photo, however, they remain vulnerable to full unlearned graph recovery. This contrast reveals that MIA-based verification, while acting as a useful first-line test, systematically underestimates the privacy risks in GU. Our findings thus call for more rigorous privacy auditing standards that integrate reconstruction-level tests.

\subsubsection{Defense (SQ8)}
To investigate the performance of defense mechanisms against GraphToxin, we use the most vulnerable and stable attack target—Retraining after single-node removal in the white-box setting. We adopt both in-processing and post-processing privacy-preserving techniques. Pre-processing techniques are out of our scope as they typically alter graphs before training, making it difficult to disentangle whether observed effects stem from unlearning or from data modification.

\noindent\textbf{Node-level Differential Privacy (Node-DP) in GNNs} Node-DP in GNNs aims to protect an individual node and all its incident edges during message passing. Compared to standard or edge-DP, enforcing node-DP is substantially more challenging due to structural coupling and multi-hop dependencies, which amplify per-sample sensitivity and complicate privacy accounting. Existing studies either inject noises to shallow GNNs \cite{1_layer_dp} or redesign architectures to decouple graph convolutions \cite{gap, dpar}. Despite these efforts, achieving strong node-level privacy guarantees with competitive utility remains an open problem \cite{dp_amplify, personal_node_dp, node_dp}. 

Following Xiang et al. \cite{node_dp}, we adopt HeterPoison to implement node-DP. As the default configuration leads to severe utility degradation on GNNs, we directly tune the noise scale $\sigma \in \{0.006, 0.007, 0.008, 0.009, 0.01\}$ from the symmetric multivariate Laplace noise $\mathcal{LAP}(0, \sigma^2\mathbb{I}^d)$, resulting in test accuracies of 79.40\%, 76.00\%, 69.90\%, 61.00\% and 51.90\%, respectively. 
From Table \ref{toxin_perf_defense}, \textit{node-DP fails to effectively mitigate GraphToxin}. At smaller noise scales, the reconstruction quality remains comparable to the undefended setting, and attack accuracy may even increase. We attribute this behavior to the injected DP noise, which suppresses weak and unstable gradient components while preserving more informative signals for recovery. In this sense, node-DP can act as a form of regularization, preventing GraphToxin from overfitting to trivial details. At larger noise scales, no meaningful privacy-utility trade-off exists, as both GraphToxin's reconstruction quality and the original model's utility significantly degrade.


\noindent\textbf{Gradient Compression (GC).} Given that GraphToxin exploits the real or surrogate gradient difference for recovery, it is imperative to enhance the privacy of released gradients. Gradient compression techniques are widely adopted against such gradient-based attacks \cite{gia_survey}. Following Vogels et al. \cite{powersgd}, we prune a proportion $\mathrm{p}$ of model gradients to assess the impact of gradient sparsity on GraphToxin. We experiment with high pruning ratios of $\mathrm{p}\in\{0.5, 0.7, 0.9\}$. The most stringent case, $\mathrm{p}=0.9$, enforces the extreme sparsity where only very few gradients are preserved. The results are shown in Table \ref{toxin_perf_defense}. Similar to node-DP, \textit{GC cannot effectively defend GraphToxin. Preserving only partial gradients can lead to attack performance comparable to or even exceeding that obtained from full gradients. Even under extreme sparsification, GraphToxin remains competitive relative to the undefended baseline}. This finding further supports our previous speculation that retaining only important and informative gradients rather than replicating trivial details can enhance the attack performance.

\begin{table}[htbp]
\centering 
\caption{\centering \textbf{The Performance of Node-DP and Gradient Compression in Defending GraphToxin (Dataset: Cora)}} 
\label{toxin_perf_defense}
\resizebox{\linewidth}{!}{
\begin{tabular}{c|c|cccccc|cccccc}
\hline
\multirow{2}{*}{\textbf{Defense}} & \multirow{2}{*}{\textbf{Type}} & \multicolumn{2}{c}{\textbf{RNMSE($\downarrow$)}} & \multicolumn{2}{c}{\textbf{ED($\downarrow$)}} & \multicolumn{2}{c|}{\textbf{PMGK($\uparrow$)}} & \multicolumn{2}{c}{\textbf{ACC($\uparrow$)}} & \multicolumn{2}{c}{\textbf{FID($\uparrow$)}} & \multicolumn{2}{c}{\textbf{PWD($\downarrow$)}} \\ \cline{3-14} 
                                  &                                & Random           & Worst           & Random          & Worst         & Random           & Worst           & Random          & Worst          & Random          & Worst          & Random          & Worst         \\ \hline
\multirow{5}{*}{\textbf{DP}}         & $\sigma\textit{=}0.006$                           & \textbf{4.94}            & \textbf{2.63}            & \textbf{65.79}           & \textbf{54.16}         & \textbf{84.80}            & 65.81           & 77.13           & \textbf{90.92}          & \textbf{66.17}           & \textbf{80.98}          & \textbf{41.64}           & \textbf{36.40}         \\ 

                                    & $\sigma\textit{=}0.007$                           & 5.88             & 3.02            & 67.95           & 61.79         & 82.39            & 65.44           & \textbf{86.72}           & 89.87          & 57.06           & 73.69          & 44.84           & 45.15         \\ 

                                    & $\sigma\textit{=}0.008$                           & 5.86             & 3.22            & 73.16           & 67.57         & 82.10            & 66.76           & 75.22           & 80.22          & 45.38           & 54.76          & 45.61           & 47.19         \\ 

                                    & $\sigma\textit{=}0.009$                           & 5.44             & 3.33            & 81.01           & 77.56         & 81.17            & \textbf{67.02}           & 50.13           & 65.97          & 31.45           & 47.03          & 55.31           & 52.44         \\ 

                                    & $\sigma\textit{=}0.01$                           & 5.15             & 3.30            & 89.03           & 82.55         & 82.63            & 65.68           & 34.38           & 52.80          &47.03           & 36.95          & 54.12           & 49.95         \\ \hline
\multirow{3}{*}{\textbf{GC}}               & p=0.5                          & 4.15             & \textbf{2.05}            & 70.40           & 48.40         & \textbf{83.81}            & 67.20           & 76.62           & \textbf{92.81}          & 68.64           & 79.26          & 31.21           & \textbf{17.50}         \\
                                  & p=0.7                          & 4.29             & 2.15            & 70.06           & 50.99         & 83.27            & 66.16           & 73.28           & 86.60          & 71.50           & \textbf{80.54}          & 26.74           & 18.44         \\
                                  & p=0.9                          & \textbf{4.14}             & 2.15            & \textbf{66.35}           & \textbf{46.17}         & 82.53            & \textbf{67.27}           & \textbf{79.53}           & 90.31          & \textbf{74.17}           & 79.60          & \textbf{24.24}           & 17.79         \\ \hline
\end{tabular}}
\end{table}

\subsubsection{Limitation and Future Work}
\hfill




\noindent\textbf{Large-scale Node Removal} This work follows existing graph unlearning research and primarily examines GraphToxin under relatively small-scale node removal. Although the worst-case scenarios generate comparatively large unlearned graphs by simultaneously removing multiple highest-degree nodes, investigating the scalability of GraphToxin under large-scale multiple-node removal presents a promising direction.

\noindent\textbf{Beyond Degree-Based Worst-case Analysis} Node degree provides a direct and interpretable proxy for structural influence in the worst-case evaluation, as deleting a node necessarily removes all of its incident edges.
Other measures such as node influence \cite{node_influence, GIF} may capture different worst-case behaviors. Exploring alternative influence criteria could provide a more comprehensive view of GU-FGRA.

\noindent\textbf{Deep Investigation on Complex Interactions} Several non-trivial findings emerge from the experimental results. For instance, GraphToxin remains effective under strict differential privacy guarantees, and retaining only a small fraction of gradients can suffice for GU-FGRA. These observations suggest complex interactions among model architectures, node homophily and other possible factors. A deeper investigation on these factors is left for future work.

\section{Conclusion}
\label{sec:conclusion}

In this study, we pioneer the investigation into the vulnerabilities of reconstructing the full unlearned graphs through GU. We propose GraphToxin, the first full graph reconstruction attack against GU. Specifically, we introduce a novel curvature matching module to provide fine-grained signals for full unlearned graph recovery. We further extend GraphToxin to multiple-node removal under both white-box and black-box settings. Furthermore, we present a comprehensive evaluation framework, demonstrating the necessity of worst-case evaluation and multi-level evaluation metrics. Extensive experiments demonstrates the effectiveness and flexibility of GraphToxin across both white-box and black-box settings for single and multiple-node removal. In addition, our assessment of existing defenses and MIA-based verifications shows that they provide limited protection and may exhibit unintended effects on this attack, and even underestimate privacy risks, highlighting the urgent need to develop more robust defenses against the attack and more rigorous privacy auditing of GU.


\appendix

\section*{APPENDIX}


\section*{Open Science}
In accordance with the CFP’s Open Science guidelines, we release the artifacts required to reproduce our experiments, including the datasets, code, implementation details, and evaluation pipeline. To preserve anonymity during review, these materials are hosted in an anonymous repository \url{https://anonymous.4open.science/r/GraphToxin-A854}, and will be migrated to a public GitHub repository in the future. This artifact release is intended to facilitate further investigation into vulnerabilities in graph unlearning and to inspire future work on robust defenses and privacy auditing.

\section*{Ethical Considerations}
This study investigates vulnerabilities of graph unlearning solely on publicly available benchmark datasets. All graph unlearning methods were tested on local devices. It does not directly compromise personal information or cause harm to real-world systems or servers. As a red team, we validate that the proposed attack reveals substantially more sensitive information than existing inference attacks and empirically confirm that existing MIA-based verifications underestimate real privacy risks. 

The intended users are corresponding practitioners, authorized auditors, and researchers evaluating unlearning guarantees before deploying models in sensitive domains or developing stronger defenses, not malicious parties seeking to attack systems without authorization. 

\section*{LLM Acknowledgements}
We use Claude only for editorial assistance in this manuscript, and all outputs were inspected by the authors to ensure accuracy and originality.

\section{Theorem Assumptions and Proof}
\label{assu_proof}
The subsequent theoretical proof of curvature matching is built upon the following two assumptions:

\begin{assumption}[\textbf{Empirical Risk Function}]
    We assume that $\mathcal{L}(\cdot)$ is twice continuously differentiable and the Hessian $H=\nabla^2\mathcal{L}(\theta)$ is invertible and symmetric w.r.t. parameter $\theta$. Following previous studies \cite{hessian,hessian_influence}, we add a damping term $\lambda I$ to $H$ if $\mathcal{L}(\cdot)$ is non-convex.
\end{assumption}

\begin{assumption}[\textbf{Small Parameter Change}]
    We assume that the parameter change $\theta^\ast-\theta$ before and after graph unlearning is fairly small. 
\end{assumption}

We empirically observe that this change is small, as reflected in our experimental results. This observation is consistent with the objective of graph unlearning and is theoretically bounded in prior work \cite{GIF,chien2022certified}.

Under the above assumptions, the first-order Taylor expansion of the gradient $\nabla\mathcal{L}(\cdot)$ around a stationary point $\theta$ is:
\begin{equation}
\label{taylor}
    \begin{split}
        \nabla\mathcal{L}(\theta^\ast)\approx\nabla\mathcal{L}(\theta)+\nabla^2\mathcal{L}(\theta)(\theta^\ast-\theta)
    \end{split}
\end{equation}

Gradient-based GU-FGRA initializes a dummy graph and iteratively optimizes it to recover a final graph whose induced gradients explain the observed parameter changes \cite{first_unlearn_attack}. Hence:
\begin{equation}
\label{gradient_descent}
    \Delta\tilde{\nabla}\approx \theta^\ast-\theta
\end{equation}
where $\Delta\tilde{\nabla}$ denotes the parameter change induced by the dummy graph. Since the original model is already trained and released, its parameters $\theta$ are fixed. Without loss of generality, GU-FGRA aims to recover parameters $\tilde{\theta}$ such that $\Delta\tilde{\nabla}=\tilde{\theta}-\theta$.

Based on Equation (\ref{taylor}) and Equation (\ref{gradient_descent}), we provide the proof of Theorem \ref{theorem_main} below:

\noindent\textit{Proof}.
Based on Equation (\ref{taylor}), we get $\Delta\nabla\approx H(\theta^\ast-\theta)$. Since H is invertible, $\theta^\ast-\theta\approx H^{-1}\Delta\nabla$. As for Equation (\ref{gradient_descent}), under the same linearized relation in Equation (\ref{taylor}), $\tilde{\theta}-\theta\approx H^{-1}\Delta\tilde{\nabla}$. Therefore, the attack goal is transformed to minimize the difference between $H^{-1}\Delta\nabla$ and $H^{-1}\Delta\tilde{\nabla}$. Formally, $H^{-1}(\Delta\tilde{\nabla}-\Delta\nabla)\approx0$. Take the $H^{1/2}$ to the left-hand and right-hand side, then we get $H^{-1/2}(\Delta\tilde{\nabla}-\Delta\nabla)\approx0$. Consider the squared L2 norm, the previous approximation equals to $(\Delta\tilde{\nabla}-\Delta\nabla)^TH^{-1}(\Delta\tilde{\nabla}-\Delta\nabla)$, hereby completing the proof.

\section{Experimental Validation}

\subsection{Datasets}
\label{data}
Cora and PubMed are small-scale and medium-scale citation networks with low average node degrees, where nodes represent scientific/medical publications, edges denote citations, and node features carry author's personal information. Photo is a dense co-purchase graph, where nodes represent products, edges indicate frequent co-purchases, and features encode product reviews. The dataset statistics are shown in Table \ref{data_stat}. We follow the standard data split setting commonly used in GNN benchmarks \cite{amazon_photo,gnn_bench_nas}, randomly selecting $20$ nodes per class for training, $30$ nodes per class for validation, and using the remaining nodes for testing. 

\begin{table}[!htbp]
\caption{\centering \textbf{Detailed Graph Statistics}}
\label{data_stat}
\resizebox{\linewidth}{!}{
\begin{tabular}{c|cccc|cc}
\toprule
\textbf{Datasets}  & \textbf{Nodes}  & \textbf{Edges}   & \textbf{Features} & \textbf{Labels} & \textbf{Avg. Deg.} & \textbf{Homo.(\%)} \\ \hline
\textbf{Cora}      & 2,708  & 10,556  & 1,433    & 7      & 3.90   &   82.52      \\
\textbf{PubMed}  & 19,717  & 88,648   & 500    & 3      & 4.50   &   79.24     \\
\textbf{Photo} & 7,650 & 238,162 & 745      & 8     & 31.13   &   83.65    \\ \bottomrule
\end{tabular}}
\end{table}

\subsection{Implementation}
\label{implement}
\begin{enumerate}
\item \textbf{Victim GNN}: We consider 2-layer GNNs with the hidden dimension of $256$ and a linear classifier for node classification tasks. We tune the learning rate from $\{0.001, 0.005\}$, and set weight decay as $5e\!-\!4$ and training epochs as $200$. 
    \item \textbf{Graph Unlearning}: For MEGU, we adopt the default setting to set the unlearning rate as $0.05$, weight decay as $1e\!-\!4$, tuning epoch as 100, and coefficients in the propagation process as $0.8$, $0.5$ and the coefficient in the loss function as $0.01$. For ETR, we follow their default setting to tune the parameter--rectifying coefficient $\lambda$ from $\{0.001, 0.01, 0.3\}$. 
    \item \textbf{GraphToxin}: For single-node removal, we randomly sample 10\% of training nodes or select the top 10\% highest-degree nodes as attack targets, and report their average results throughout our paper. The attacker does not have access to the locations of the removed nodes and only observes the gradient difference between the original and the unlearned GNNs. Unless otherwise specified, the Fisher coefficient is set to $1$ and the Laplacian coefficient is set to $1e\!-\!5$. AdamW \cite{adamw} is used as the optimizer. Since multiple restarts to the optimizer (e.g, L-BFGS) from different starting points can significantly improve the reconstruction quality, for a fair comparison, we strictly follow the configuration in Sinha et.al. \cite{gradient_attack_gnn} for all experiments and only report the attack results with a single run of the optimizer. In addition, we strictly set the attack learning rate as $0.01$ for all experiments. 
    Following Geiping et al. \cite{first_gradient} and Hu et al. \cite{first_unlearn_attack}, each optimization is run for up to $10,000$ iterations, with most attacks converging earlier, and a step size decay is applied after $3/8$, $5/8$, $7/8$ of total iterations, using the multiplicative factor of $0.5$. 
    \item \textbf{GLG-GU}: We follow the default parameter settings in GLG \cite{gradient_attack_gnn} and set the feature smoothness as $1e\!-\!9$ and the sparsity regularizer as $1e\!-\!7$. The rest optimizer, learning rate and training epochs are the same as GraphToxin.
\end{enumerate}

\subsection{Original and Unlearning Performance} 
\label{origin}
Table \ref{data_perf} shows that all graph unlearning methods exhibit comparable performance across the three GNN backbones, with variations within $1\%$. As for GNN backbones, the GCN backbone achieves the highest performance on Cora and PubMed while SGC performs best on Photo, with differences of less than 2\% compared to GCN. Across unlearning methods, ETR preserves model utility more effectively on Cora and PubMed, while Retraining exceeds other unlearning methods with GCN and SGC backbones. In contrast, MEGU yields the strongest results with GraphSAGE on Photo. Regarding removal settings, removing the $10\%$ of the highest-degree nodes results in larger benign performance degradation compared to the random-node removal, which is amplified on sparse and smaller-scale graphs.

\begin{table*}[htbp]
\caption{\centering \textbf{The Performance of the Original and Unlearned GNNs for Single-node Removal}.}
\label{data_perf}
\resizebox{\linewidth}{!}{
\begin{tabular}{c|c|cc|cccc|cccc|cccc}
\toprule
\multirow{3}{*}{\textbf{Dataset}} & \multirow{3}{*}{\textbf{GNN}} & \multicolumn{2}{c|}{\textbf{Original}}      & \multicolumn{4}{c|}{\textbf{Retraining}}                        & \multicolumn{4}{c|}{\textbf{MEGU}}                          & \multicolumn{4}{c}{\textbf{ETR}}                          \\ \cline{3-16} 
                                  &                               & \multirow{2}{*}{ACC} & \multirow{2}{*}{AUC} & \multicolumn{2}{c|}{Random}    & \multicolumn{2}{c|}{Worst} & \multicolumn{2}{c|}{Random}    & \multicolumn{2}{c|}{Worst} & \multicolumn{2}{c|}{Random}    & \multicolumn{2}{c}{Worst} \\ \cline{5-16} 
                                  &                               &                      &                      & ACC & \multicolumn{1}{c|}{AUC} & ACC          & AUC         & ACC & \multicolumn{1}{c|}{AUC} & ACC          & AUC         & ACC & \multicolumn{1}{c|}{AUC} & ACC         & AUC         \\ \hline
\multirow{3}{*}{\textbf{Cora}}    & GCN                           &80.20                      &\textbf{96.70}                      &80.05     & \multicolumn{1}{c|}{\textbf{96.68}}    &79.99              &\textbf{96.67}             &80.06     & \multicolumn{1}{c|}{\textbf{96.66}}    &\textbf{80.52}              &\textbf{96.69}             &80.57     & \multicolumn{1}{c|}{\textbf{96.76}}    &80.69             &\textbf{96.77}             \\
                                  & SAGE                          &\textbf{81.10}                      &96.61                      &\textbf{80.99}     & \multicolumn{1}{c|}{96.66}    &\textbf{80.77}              &96.64             &80.04     & \multicolumn{1}{c|}{96.54}    &79.94              &96.52             &80.80     & \multicolumn{1}{c|}{96.56}    &80.86             &96.57             \\
                                  & SGC                           &\textbf{81.10}                      &96.12                      &80.42     & \multicolumn{1}{c|}{96.24}    &80.24              & 96.19            &\textbf{80.64}     & \multicolumn{1}{c|}{96.06}    &80.49              &96.06             &\textbf{81.10}     & \multicolumn{1}{c|}{96.15}    &\textbf{81.10}             &96.17             \\ \hline
\multirow{3}{*}{\textbf{PubMed}}  & GCN                           &77.40                      &\textbf{90.44}                      &77.78     & \multicolumn{1}{c|}{\textbf{90.53}}    &\textbf{77.93}              &\textbf{90.40}             &\textbf{77.12}     & \multicolumn{1}{c|}{\textbf{90.28}}    &\textbf{76.68}              &\textbf{90.32}             &77.53     & \multicolumn{1}{c|}{\textbf{90.45}}    &77.75             &\textbf{90.46}             \\
                                  & SAGE                          &76.90                      &89.44                      &75.53     & \multicolumn{1}{c|}{89.23}    &74.80              &89.08             &76.22     & \multicolumn{1}{c|}{89.51}    &76.05              &89.50             &76.68     & \multicolumn{1}{c|}{89.54}    &76.48             &89.53             \\
                                  & SGC                           &\textbf{78.20}                      &90.37                      &\textbf{77.92}     & \multicolumn{1}{c|}{90.44}    &77.78              &90.29             &75.25     & \multicolumn{1}{c|}{89.65}    &75,72              &90.03             &\textbf{78.05}     & \multicolumn{1}{c|}{89.90}    &\textbf{78.12}             &89.95             \\ \hline
\multirow{3}{*}{\textbf{Photo}}   & GCN                           &90.07                      &99.06                      &91.09     & \multicolumn{1}{c|}{\textbf{99.05}}    &91.03              &99.06             &89.11     & \multicolumn{1}{c|}{98.90}    &89.47              &98.96             &88.19     & \multicolumn{1}{c|}{98.71}    &88.26             &98.72             \\
                                  & SAGE                          &91.31                      &\textbf{99.28}                      &89.22     & \multicolumn{1}{c|}{\textbf{99.05}}    &89.86              &\textbf{99.09}             &90.40     & \multicolumn{1}{c|}{\textbf{99.14}}    &90.15              &\textbf{99.14}             &80.45     & \multicolumn{1}{c|}{93.14}    &80.38             &93.13             \\
                                  & SGC                           &\textbf{92.04}                      &99.02                      &\textbf{91.14}     & \multicolumn{1}{c|}{98.99}    &\textbf{91.40}              &99.00             &\textbf{91.08}    & \multicolumn{1}{c|}{98.88}    &\textbf{90.97}              &98.94             &\textbf{90.11}     & \multicolumn{1}{c|}{\textbf{99.03}}    &\textbf{90.12}             &\textbf{99.03}             \\ \bottomrule
\end{tabular}}
\end{table*}

\subsection{Main Attack Performance}
\begin{table*}[htbp]
\caption{\centering \textbf{The Performance of GraphToxin for Single-node Removal (Backbone: GCN)}. Cont.}
\label{toxin_perf_app}
\resizebox{\linewidth}{!}{
\begin{tabular}{c|c|c|cccccc|cccccc}
\toprule
\multirow{2}{*}{\textbf{Dataset}} & \multirow{2}{*}{\textbf{GU}} & \multirow{2}{*}{\textbf{Baseline}} & \multicolumn{2}{c}{\textbf{RNMSE($\downarrow$)}} & \multicolumn{2}{c}{\textbf{ED($\downarrow$)}} & \multicolumn{2}{c|}{\textbf{PMGK($\uparrow$)}} & \multicolumn{2}{c}{\textbf{ATT. ACC($\uparrow$)}} & \multicolumn{2}{c}{\textbf{ATT. FID($\uparrow$)}} & \multicolumn{2}{c}{\textbf{PWD($\downarrow$)}} \\ \cline{4-15} 
                                  &                              &                                    & Random           & Worst           & Random          & Worst         & Random           & Worst           & Random          & Worst          & Random          & Worst          & Random          & Worst         \\ \hline
\multirow{9}{*}{\textbf{PubMed}}           & \multirow{3}{*}{Retrain}       & Rand.                              &79.58                  &28.85                 &68.92                 &70.95               &91.62                  &68.36                 &67.46                 &29.76                &74.07                 &48.19                &107.87                 &\textcolor{blue}{\textbf{2.59}}               \\
                                  &                              & FewE                               &77.73                  &27.99                 &\textbf{65.14}                 &\textbf{61.47}               &\textbf{92.30}                  &\textbf{71.30}                 &67.46                 &77.19                &74.08                 &67.20                &\textcolor{blue}{\textbf{0.02}}                 &10.61               \\
                                  &                              & GLG-GU  &\textbf{45.74} &\textbf{10.04} &68.59 &90.81 &90.47 &64.55 &68.39 &38.67 &77.78 &30.99 &20.89 &61.74
   \\
                                  &                              & Toxin                              &53.44                  &13.36                 &74.17                 &63.16               &90.66                  &70.68                 &\textbf{85.58}                 &\textcolor{blue}{\textbf{88.18}}                &\textbf{78.17}                 &\textcolor{blue}{\textbf{69.14}}               &18.65                 &12.24               \\ \cline{2-15} 
                                  & \multirow{3}{*}{MEGU}        & Rand.                              &79.58                  &28.85                 &68.92                 &70.95               &92.06                  &70.05                 &67.46                 &29.76                &74.07                 &48.19                &107.87                 &\textcolor{blue}{\textbf{2.59}}               \\
                                  &                              & FewE                               &77.73                  &27.99                 &\textbf{65.14}                 &\textbf{61.47}               &\textbf{92.55}                  &\textbf{71.16}                 &67.46                 &77.19                &74.07                 &67.20                &\textcolor{blue}{\textbf{0.02}}                 &10.61               \\
                                  &                              & GLG-GU  &\textbf{45.74} &\textbf{10.00} &68.59 &90.81 &90.46 &64.31 &68.39 &38.67 &77.78 &30.99 &20.89 &61.74   \\
                                  &                              & Toxin                              &53.44                  &13.36                 &74.17                 &63.16               &91.17                  &70.42                 &\textbf{85.58}                 &\textcolor{blue}{\textbf{88.18}}                & \textbf{78.17}                &\textcolor{blue}{\textbf{69.14}}                &\textbf{18.65}                 &\textbf{12.24}               \\ \cline{2-15} 
                                  & \multirow{3}{*}{ETR}         & Rand.                              &79.58                  &28.85                 &68.92                 &70.95               &91.59                  &69.17                 &67.46                 &29.76                &74.07                 &48.19                &107.87                 &\textcolor{blue}{\textbf{2.59}}               \\
                                  &                              & FewE                               &77.73                  &27.99                 &\textbf{65.14}                 &\textbf{61.47}               &\textbf{92.61}                  &\textbf{72.63}                 &67.46                 &77.19                &74.08                 &67.20                &\textcolor{blue}{\textbf{0.02}}                 &10.61               \\
                                  &                              & GLG-GU  &\textbf{45.74} &\textbf{10.00} &68.59 &90.81 &90.72 &65.09 &68.39 &38.67 &77.78 &30.99 &20.89 &61.74   \\
                                  &                              & Toxin                              &53.44                  &13.36                 &74.17                 &63.16               &91.53                  &69.05                 &\textbf{85.58}                 &\textcolor{blue}{\textbf{88.18}}                &\textbf{78.17}                 &\textcolor{blue}{\textbf{69.14}}                &\textbf{18.65}                 &\textbf{12.24}                        \\ \bottomrule
\end{tabular}}
\end{table*}

\subsection{Attack with GNN Backbones}
As shown in Table \ref{toxin_perf_backbone}, \textit{GraphToxin consistently retains its superiority and generalization across all GNN backbones and graph datasets}. Taken together, these results suggest that there is no universally most vulnerable GNN due to the complex interactions among model architecture, graph sparsity and homophily. We provide the following plausible explanations: (1) GCN aggregates information over local neighborhoods, which makes node representations strongly dependent on features propagated from their local neighborhoods. This propagation can provide more informative gradient signals for recovering underlying graph structure. However, on dense graphs, rapid neighborhood expansion can lead to over-smoothing, weakening node-specific signals. As a result, GCN tends to achieve more competitive performance on small-scale and sparse graphs such as Cora. (2) GraphSAGE relies on random neighbor sampling rather than full neighborhood aggregation. Consequently, gradients are computed from incomplete and randomly varying local structures, which might obfuscate the gradient-based guidance for recovery, particularly on sparse graphs. While on dense graphs such as Photo, structural redundancy can offset the randomness introduced by neighbor sampling, enabling GraphToxin with the GraphSAGE backbone to maintain strong reconstruction performance. (3) SGC employs linear multi-hop feature propagation. Although its linearity offers a more direct and simpler proxy for full graph recovery, it can suppress fine-grained and node-specific details. This dual effect might explain why SGC performs well in some metrics on PubMed and Photo, yet exhibits degraded performance on Cora.
\label{att_backbone}
\begin{table*}[htbp]
\caption{\centering \textbf{GraphToxin for Single-node Removal with Different GNN Backbones (GU: Retraining)}.}
\label{toxin_perf_backbone}
\resizebox{\linewidth}{!}{
\begin{tabular}{c|c|c|cccccc|cccccc}
\toprule
\multirow{2}{*}{\textbf{Dataset}} & \multirow{2}{*}{\textbf{GNN}} & \multirow{2}{*}{\textbf{Baseline}} & \multicolumn{2}{c}{\textbf{NRMSE($\downarrow$)}} & \multicolumn{2}{c}{\textbf{ED($\downarrow$)}} & \multicolumn{2}{c|}{\textbf{PMGK($\uparrow$)}} & \multicolumn{2}{c}{\textbf{ATT. ACC($\uparrow$)}} & \multicolumn{2}{c}{\textbf{ATT. FID($\uparrow$)}} & \multicolumn{2}{c}{\textbf{PWD($\downarrow$)}} \\ \cline{4-15} 
                                  &                               &                                    & Random           & Worst           & Random          & Worst         & Random           & Worst           & Random          & Worst          & Random          & Worst          & Random          & Worst         \\ \hline
\multirow{9}{*}{\textbf{Cora}}    & \multirow{3}{*}{GCN}          & Rand.                              &6.59                  &3.61                 &78.76                 &72.45               &84.35                  &68.35                 &6.79                 &9.18                &6.31                 &7.48                &82.79                 &80.23               \\
                                  &                              & FewE                               &5.08                  &2.84                 &73.11                 &65.96               &\textbf{84.62}                  &\textbf{68.79}                 &57.50                 &39.29                &56.67                 &37.98                &40.47                 &56.30               \\
                                  &                              & GLG-GU  &\textbf{3.82} &1.98 &96.83 &71.57 &79.74 &66.38 &28.33 &45.44 &25.71 &40.73 &68.54 &53.66   \\
                                  &                              & Toxin                              &4.11                  &\textbf{1.95}                 &\textbf{71.38}                 &\textbf{52.35}               &84.05                  &67.21                 &\textcolor{blue}{\textbf{72.21}}                 &\textcolor{blue}{\textbf{84.72}}                &\textcolor{blue}{\textbf{69.00}}                 &\textcolor{blue}{\textbf{77.97}}                &\textcolor{blue}{\textbf{29.66}}                 &\textcolor{blue}{\textbf{22.35}} \\ \cline{2-15} 
                                  & \multirow{3}{*}{SAGE}         & Rand.                              &6.59                  &3.61                 &65.84                 &63.10               &81.13                  &66.78                 &6.79                 &9.18                &5.60                 &6.66                &89.97                 &91.17               \\
                                  &                               & FewE                               &6.01                  &3.24                 &67.71                 &64.14               &80.10                  &66.94                 &38.69                 &28.63                &33.93                 &24.36                &56.19                 &65.42               \\
                                  &                              & GLG-GU  &\textbf{2.50} &\textbf{1.32} &88.72 &83.00 &79.23 &66.65 &28.93 &33.23 &23.57 &24.43 &72.89 &73.88   \\
                                  &                               & Toxin                              &4.49                  &1.86                 &\textbf{60.20}                 &\textbf{57.46}               &\textbf{82.02}                  &\textbf{67.10}                 &\textcolor{blue}{\textbf{68.21}}                 &\textcolor{blue}{\textbf{64.36}}                &\textcolor{blue}{\textbf{62.26}}                 & \textcolor{blue}{\textbf{55.17}}               &\textcolor{blue}{\textbf{33.93}}                 &\textcolor{blue}{\textbf{38.72}}               \\ \cline{2-15} 
                                  & \multirow{3}{*}{SGC}          & Rand.                              &6.59                  &3.61                 &95.25                 &91.71               &81.41                  &67.04                 &6.79                 &9.18                &8.10                 &9.26                &84.98                 &80.48               \\
                                  &                               & FewE                               &6.02                  &3.38                 &79.64                 &80.47               &80.32                  &67.60                 &32.14                 &28.66                &32.86                 &28.76                &63.28                 &68.44               \\
                                  &                              & GLG-GU  &\textbf{2.58} &\textbf{1.54} &107.00 &109.46 & 77.57&65.86 &18.93 &20.32 &18.81 &19.17 &80.45 &76.00   \\
                                  &                               & Toxin                              &4.35                  &2.16                 &\textbf{65.47}                 &\textbf{66.94}               &\textbf{84.04}                  &\textbf{67.56}                 &\textcolor{blue}{\textbf{72.50}}                 &\textcolor{blue}{\textbf{61.03}}                &\textcolor{blue}{\textbf{67.50}}                 &\textcolor{blue}{\textbf{55.99}}                &\textcolor{blue}{\textbf{30.97}}                 &\textcolor{blue}{\textbf{41.89}}               \\ \hline
\multirow{9}{*}{\textbf{PubMed}}  & \multirow{3}{*}{GCN}          & Rand.                              &79.58                  &28.85                 &68.92                 &70.95               &91.62                  &68.36                 &67.46                 &29.76                &74.07                 &48.19                &107.87                 &\textcolor{blue}{\textbf{2.59}}               \\
                                  &                              & FewE                               &77.73                  &27.99                 &\textbf{65.14}                 &\textbf{61.47}               &\textbf{92.30}                  &\textbf{71.30}                 &67.46                 &77.19                &74.08                 &67.20                &\textcolor{blue}{\textbf{0.02}}                 &10.61               \\
                                  &                              & GLG-GU  &45.74 &100.04 &68.59 &90.81 &90.47 &64.55 &68.39 &38.67 &77.78 &30.99 &20.89 &61.74   \\
                                  &                              & Toxin                              &\textcolor{blue}{\textbf{53.44}}                  &\textcolor{blue}{\textbf{13.36}}                 &74.17                 &63.16               &90.66                  &70.68                 &\textbf{85.58}                 &\textcolor{blue}{\textbf{88.18}}                &\textbf{78.17}                 &\textcolor{blue}{\textbf{69.14}}               &18.65                 &12.24               \\ \cline{2-15} 
                                  & \multirow{3}{*}{SAGE}         & Rand.                              &79.58                  &28.85                 &125.38                 &99.69               & 91.28                 &\textbf{72.51}                 &16.80                 &27.29                &1.85                 & 33.33               &59.51                 &50.23               \\
                                  &                               & FewE                               &79.12                  &28.03                 &86.45                 &74.73               &\textbf{93.24}                  &69.55                 &65.21                 &64.13                &55.82                 &64.16                &25.55                 &32.54               \\
                                  &                              & GLG-GU  &\textcolor{blue}{\textbf{5.09}} &\textbf{3.35} &100.70 &\textbf{67.61} &92.82 &69.17 &53.57 &58.13 &33.07 &60.08 &70.37 &30.18   \\
                                  &                               & Toxin                              &56.36                  &8.64                 &\textbf{80.12}                 &76.06               &91.32                  &60.54                 &\textcolor{blue}{\textbf{75.93}}                 &\textcolor{blue}{\textbf{84.17}}                &\textcolor{blue}{\textbf{66.53}}                 &\textcolor{blue}{\textbf{73.65}}                &\textcolor{blue}{\textbf{26.97}}                 &\textcolor{blue}{\textbf{21.28}}               \\ \cline{2-15} 
                                  & \multirow{3}{*}{SGC}          & Rand.                              &79.58                  &28.85                 &81.82                 &83.96               &91.55                  &69.86                 &67.46                 &29.76                &73.15                 &46.39                &\textbf{0.00}                 &\textbf{2.90}               \\
                                  &                               & FewE                               &78.89                  &28.79                 &67.72                 &88.19               &\textbf{91.99}                  &\textbf{69.92}                 &63.10                 &70.08                &58.33                 &53.30                &22.43                 &19.82               \\
                                  &                              & GLG-GU  &\textbf{39.96} &\textbf{6.46} &75.27 &\textbf{63.20} &89.33 &68.64 &64.68 &62.38 &66.67 &55.29 &22.35 &37.28   \\
                                  &                               & Toxin                              &42.97                  &14.55                 &\textbf{64.16}                 &70.67               &88.98                  &68.42                 & \textbf{81.88}                &\textcolor{blue}{\textbf{92.48}}                &\textbf{81.88}                 &\textbf{69.40}                &2.01                 &12.28               \\ \hline
\multirow{9}{*}{\textbf{Photo}}   & \multirow{3}{*}{GCN}          & Rand.                              &\textcolor{blue}{\textbf{0.61}}                  &\textcolor{blue}{\textbf{0.31}}                 &92.58                 &107.83               &\textbf{69.33}                  &\textbf{63.29}                 &26.01                 &2.98                &25.20                 &0.93                &74.48                 &92.92               \\
                                  &                              & FewE                               &\textcolor{blue}{\textbf{0.61}}                  &\textcolor{blue}{\textbf{0.31}}                 &51.46                 & 68.21              &68.03                  &58.10                 & 67.44                &28.76                &68.02                 &28.74                &28.12                 &62.26               \\
                                  &                              & GLG-GU  &0.70 &0.35 &66.79 &66.96 &67.20 &53.40 &51.03 &58.09 &50.39 &56.04 &47.93 &48.42   \\
                                  &                              & Toxin                              &0.64                  &0.34                 &\textcolor{blue}{\textbf{44.49}}                 &\textcolor{blue}{\textbf{52.11}}               &66.33                  &49.65                 &\textcolor{blue}{\textbf{86.11}}                 &\textcolor{blue}{\textbf{85.95}}                &\textcolor{blue}{\textbf{85.65}}                 &\textcolor{blue}{\textbf{80.85}}                &\textcolor{blue}{\textbf{13.91}}                 &\textcolor{blue}{\textbf{30.36}}               \\ \cline{2-15} 
                                  & \multirow{3}{*}{SAGE}         & Rand.                              &\textbf{0.61}                  &\textbf{0.31}                 &116.19                 &127.18               &\textbf{70.49}                  &\textbf{65.60}                 &26.01                 &2.98                &26.56                 &3.89                &74.77                 &95.12               \\
                                  &                               & FewE                                   &0.62                  &0.33                 &72.00                 &100.15               &65.23                  &56.95                 &65.29                 &29.26                &66.60                 &29.32                &29.88                 &64.59               \\
                                  &                              & GLG-GU  &0.68 &0.36 &78.36 &90.81 &65.77 &49.81 &70.46 &53.38 &69.51 &50.11 &29.95 &54.12   \\
                                  &                               & Toxin                                    &0.65                  &0.34                 &\textcolor{blue}{\textbf{55.44}}                 &\textbf{70.72}               &62.42                  &48.22                 &\textcolor{blue}{\textbf{94.52}}                 &\textcolor{blue}{\textbf{87.79}}                &\textcolor{blue}{\textbf{91.74}}                 &\textcolor{blue}{\textbf{81.67}}                &\textcolor{blue}{\textbf{11.51}}                 &\textcolor{blue}{\textbf{30.41}}               \\ \cline{2-15} 
                                  & \multirow{3}{*}{SGC}          & Rand.                              &\textbf{0.61}                  &\textbf{0.31}                 &94.57                 &111.16               &\textbf{71.23}                  &\textbf{63.74}                 &26.01                 &2.98                &25.20                 &1.25                &74.11                 &92.11               \\
                                  &                               & FewE                               &\textbf{0.61}                  &\textbf{0.31}                 &58.26                 &56.17               &68.42                  &60.97                 &64.69                 &45.54                &65.99                 &47.94                &26.50                 &41.42               \\
                                  &                              & GLG-GU  &0.73 &0.37 &77.80 &73.00 &61.95 &45.84 &53.02 &56.37 &50.17 &52.38 &48.82 &55.87   \\
                                  &                               & Toxin                              &0.66                  &0.33                 &\textbf{56.02}                 &\textcolor{blue}{\textbf{47.60}}               &65.29                  &45.59                 &\textcolor{blue}{\textbf{80.78}}                 &\textcolor{blue}{\textbf{90.39}}                &\textcolor{blue}{\textbf{76.55}}                 &\textcolor{blue}{\textbf{85.76}}                &\textcolor{blue}{\textbf{21.05}}                 &\textcolor{blue}{\textbf{26.89}}               \\ \bottomrule
\end{tabular}}
\end{table*}

\subsection{Impact Factor - Unlearned Graph Size}
\label{imp_graph_size}
\begin{figure*}[htbp]
    \centering
    \subfloat{\includegraphics[width=0.5\linewidth]{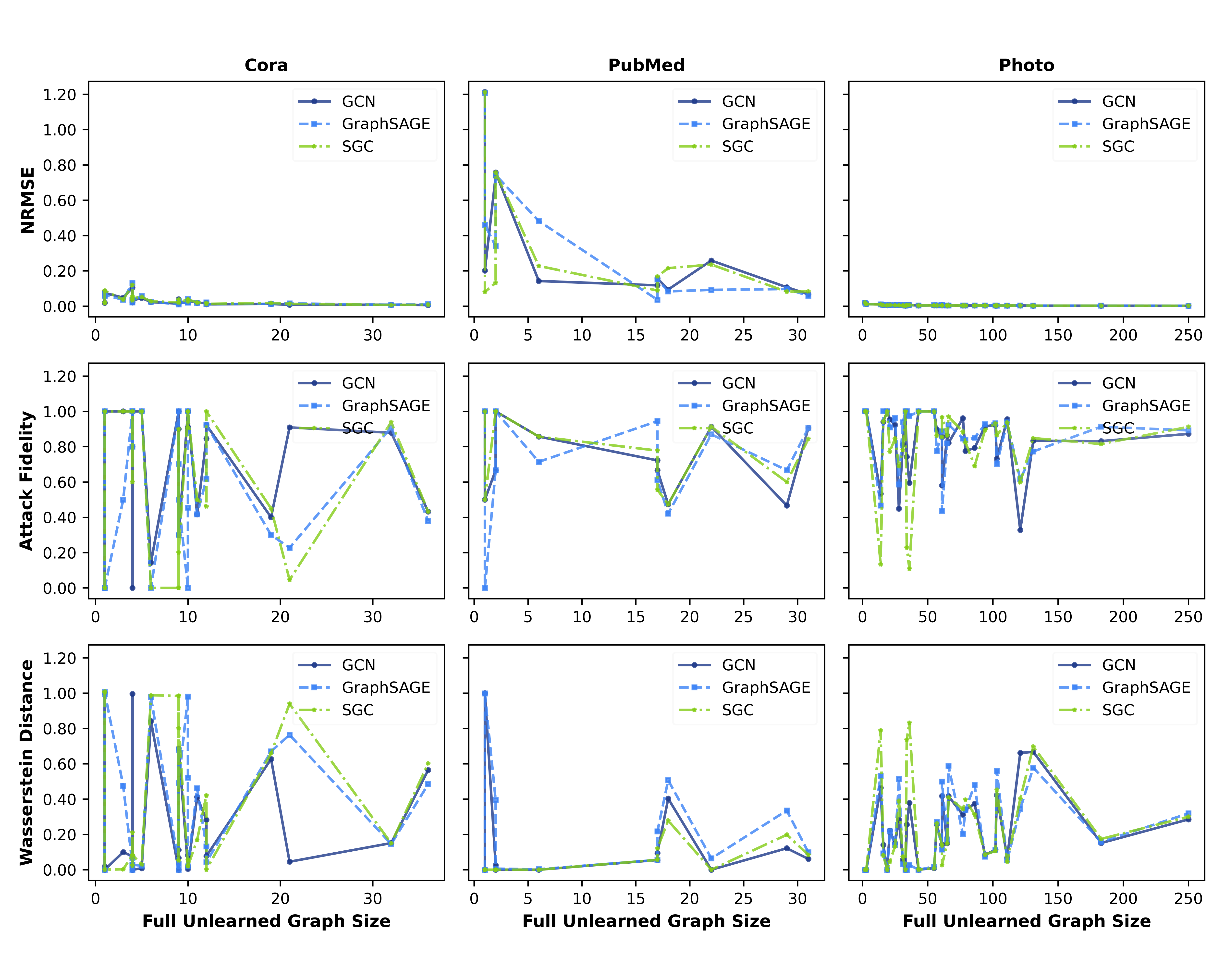}}
    \subfloat{\includegraphics[width=0.5\linewidth]{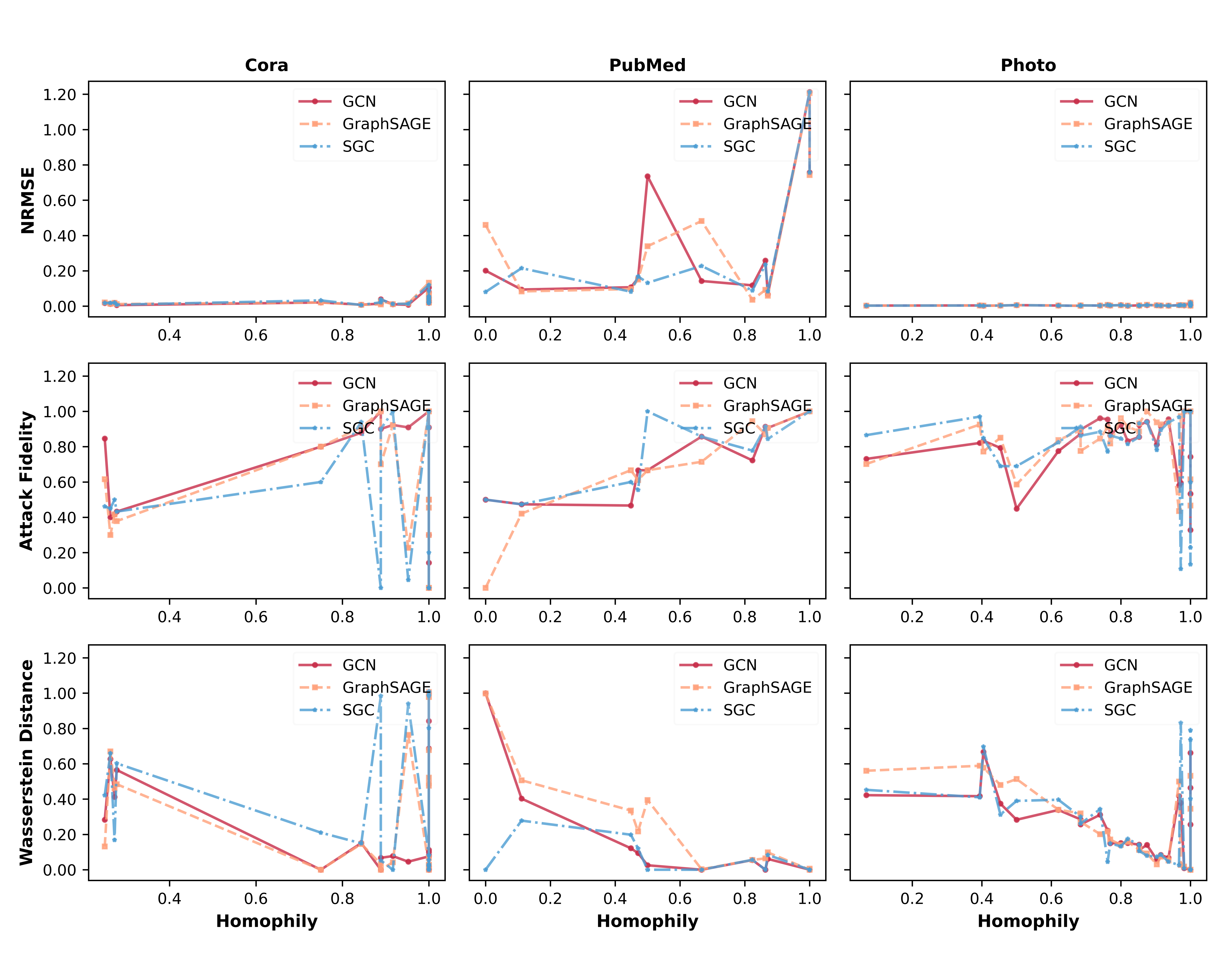}}
    \caption{Impact of the Full Unlearned Graph Size and Node Homophily}
    \label{graph_size_rest}
\end{figure*}

From Figures \ref{node_size_homo} and \ref{graph_size_rest}, RNMSE remains consistently low with minimal variation, indicating its insensitivity to the unlearned graph size. PMGK generally decreases as the graph size increases. An exception is observed on Photo with the GCN backbone, where PMGK rebounds when the deletion graph becomes sufficiently large. ED and PWD also exhibit a decreasing trend with increasing unlearned graph size. In addition, attack accuracy and fidelity show noticeable oscillations. These variations arise from single-node removal, where each target node induces an unlearned graph of different sizes and structural complexity. The results are then sorted by the size of the unlearned graph. Heterogeneity across target nodes leads to varying reconstruction difficulty, which manifests as fluctuations in the plotted results.

\subsection{Impact Factor - Node Homophily}
\label{imp_node_homo}
Consistent with the analysis on unlearned graph size, RNMSE remains largely invariant to changes in the homophily of unlearned nodes. However, other metrics exhibit clearer correlations with node homophily: PMGK as well as attack accuracy and fidelity are positively correlated with homophily, whereas ED and PWD are negatively correlated. Overall, these observations \textit{indicate that node homophily may serve as a more informative indicator than unlearned graph size for characterizing the vulnerability of GU-FGRA}. We leave the deeper investigation for future work.


\subsection{Multiple Node Removal} 
\label{multi_node}
In this section, we examine the effectiveness of GraphToxin under multiple-node removal, where $5$, $10$, and $15$ nodes are simultaneously removed from the training graph, respectively. For a fair comparison, all hyperparameters are kept identical to those in single-node removal. Under the random removal setting, we sample $5$ disjoint node sets with the same removal size, while in the worst-case removal setting, we select the top $25,50,75$ highest-degree nodes and partition them into five non-overlapping sets based on their degree ranking. The average results are reported in Table \ref{toxin_perf_multi}. 

Under the worst-case multiple-node removal, the unlearning performance of Retraining improves in accuracy but slightly declines in AUC when more nodes are removed. In terms of attack performance, GraphToxin consistently performs comparably to or better than other baselines. As the number of removed nodes increases, reconstruction-related metrics generally yield better results. In contrast, performance-level metrics display a non-monotonic pattern, which initially increase and then slightly degrade. Additionally, the random-case and worst-case scenarios exhibit complementary strengths, with neither consistently dominating the other across all metrics. This observation \textit{highlights the necessity of considering both scenarios for a comprehensive evaluation}.


\begin{table*}[!htbp]
\centering 
\caption{\centering \textbf{GraphToxin for Multiple Node Removals Based on Retraining (Backbone: GCN, Dataset: Cora)}.}
\label{toxin_perf_multi}
\resizebox{\linewidth}{!}{
\begin{tabular}{c|cc|c|cccccc|cccccc}
\toprule
\multirow{2}{*}{\textbf{Rem. Num}} & \multicolumn{2}{c|}{\textbf{Retraining(ACC/AUC)}} & \multirow{2}{*}{\textbf{Baseline}} & \multicolumn{2}{c}{\textbf{NRMSE($\downarrow$)}} & \multicolumn{2}{c}{\textbf{ED($\downarrow$)}} & \multicolumn{2}{c|}{\textbf{PMGK($\uparrow$)}} & \multicolumn{2}{c}{\textbf{ACC($\uparrow$)}} & \multicolumn{2}{c}{\textbf{FID($\uparrow$)}} & \multicolumn{2}{c}{\textbf{PWD($\downarrow$)}} \\
\cmidrule(lr){2-3} \cmidrule(lr){5-6} \cmidrule(lr){7-8} \cmidrule(lr){9-10} \cmidrule(lr){11-12} \cmidrule(lr){13-14} \cmidrule(lr){15-16}
& Random & Worst & & Random & Worst & Random & Worst & Random & Worst & Random & Worst & Random & Worst & Random & Worst \\
\midrule
\multirow{3}{*}{\textbf{5}} & \multirow{3}{*}{80.22/96.64} & \multirow{3}{*}{80.30/96.66} & Rand. &2.34 &1.71 &73.16 &73.52 &68.01 &62.14 &27.88 &9.94 &23.80 &8.68 &67.08 &77.68 \\ 
& & & FewE &2.13 &1.65 &61.95 &62.41 &\textbf{70.47} &60.59 &28.75 &20.02 &28.78 &20.15 &66.79 &72.13 \\
&                              &  & GLG-GU  &\textbf{0.91} &\textbf{0.70} &\textbf{53.86} &\textbf{52.17} &69.55 &62.78 &\textbf{68.27} &68.07 &\textbf{59.00} &61.92 &\textbf{40.57} &\textbf{38.44}   \\
& & & Toxin &1.66 &1.03 &56.27 &54.59 &70.45 &\textbf{64.82} &60.03 &\textbf{70.40} &52.06 &\textbf{63.98} &47.24 &39.08 \\
\midrule
\multirow{3}{*}{\textbf{10}} & \multirow{3}{*}{79.98/96.50} & \multirow{3}{*}{80.44/96.49} & Rand. &1.69 &1.42 &72.50 &73.56 &58.88 &63.76 &24.84 &17.72 &22.04 &16.00 &66.75 &72.57 \\
& & & FewE &1.58 &1.31 &60.10 &59.51 &64.65 &68.37 &24.80 &20.54 &23.63 &17.99 &65.72 &68.30 \\
&                              &  & GLG-GU  &\textbf{0.78} &\textbf{0.53} &\textbf{51.18} &\textbf{50.68} &\textbf{69.00} &67.05 &62.66 &57.84 &58.25 &53.30 &43.67 &49.85    \\
& & & Toxin &0.99 &0.73 &54.36 &52.29 &65.78 &\textbf{68.64} &\textbf{67.83} &\textbf{64.47} &\textbf{59.12} &\textbf{58.99} &\textbf{39.79} &\textbf{44.09} \\
\midrule
\multirow{3}{*}{\textbf{15}} & \multirow{3}{*}{79.46/96.57} & \multirow{3}{*}{79.84/96.47} & Rand. &1.41 &1.28 &73.37 &74.06 &63.11 &62.27 &21.28 &16.61 &19.87 &15.21 &70.53 &75.92 \\
& & & FewE &1.32 &1.18 &57.73 &58.69 &\textbf{70.05} &69.47 &22.71 &17.38 &21.82 &16.31 &67.23 &72.32 \\
&                              &  & GLG-GU  &\textbf{0.70} &\textbf{0.57} &\textbf{50.91} &\textbf{50.74} &67.89 &\textbf{71.66} &60.49 &57.55 &50.93 &53.28 &52.36 &53.24   \\
& & & Toxin &0.80 &0.69 &54.06 &55.67 &66.05 &70.34 &\textbf{65.63} &\textbf{58.79} &\textbf{53.80} &\textbf{53.50} &\textbf{51.01} &\textbf{51.05} \\
\bottomrule
\end{tabular}
}
\end{table*}

\subsection{Black-box Setting} 
\label{black_box_toxin}
\noindent\textbf{Setup}
We now assess the performance of GraphToxin under a data-free, black-box setting. Following the configuration in Zhuang et al. \cite{data_free_msa_gnn}, we utilize a GCN with $2$ layers and $256$ hidden units as a surrogate model, and a $3$-layer MLP with $128$ and $256$ hidden units as a full-parameter graph generator. We set the query graph size as $250$ with $32$ feature dimensions. During model stealing, each query encompasses training the generator $2$ times with a learning rate of $1e\!-\!6$ and training the surrogate $5$ times with a learning rate of $0.001$. The total number of queries is set to $100$. The stealing performance of the original GNN is $79.40\%$ in accuracy and $96.12\%$ in AUC. After unlearning, the stealing performance of the unlearned GNN under both single and multiple-node removal improves in accuracy while slightly declines in AUC. As for GU-FGRA, we adopt the same hyperparameter settings as in the white-box experiments. For the semantic calibration module, we set its coefficient as $50$ for single-node removal and $5000$ for multiple-node removal, as the latter introduces substantially higher noise. Average results are reported in Table \ref{toxin_perf_black}. 

\noindent\textbf{Single Node Removal} We observe that \textit{black-box GraphToxin consistently outperforms all baselines under single-node removal}. Since we introduce the semantic calibration module to enforce the label alignment, the overall performance in performance-level metrics is competitive, although the performance in reconstruction-related metrics degrades relative to the white-box setting as expected. A plausible explanation is that black-box GraphToxin utilizes the semantic calibration during recovery, which might shift some nodes across the model's decision boundary to obtain the correct predictions but ignore the actual distance in feature or embedding space. Overall, the degradation in such metrics remains acceptable. 

\noindent\textbf{Multiple Node Removal} \textit{Under the black-box multi-node removal setting, GraphToxin still exceeds other baselines}. Compared to single-node removal, performance-level metrics decline when more nodes are unlearned. However, the attack results of the reconstruction-related metrics are mixed. PMGK and ED decrease, while NRMSE improves in both random and worst-case node removal scenarios.

\begin{table*}[!htbp]
\centering 
\caption{\centering \textbf{Black-box GraphToxin for A Single and Multiple Node Removals Based on Retraining (Dataset: Cora)}.}
\label{toxin_perf_black}
\resizebox{\linewidth}{!}{
\begin{tabular}{c|cc|c|cccccc|cccccc}
\hline
\multirow{2}{*}{\textbf{Type}} & \multicolumn{2}{c|}{\textbf{Retraining(ACC/AUC)}} & \multirow{2}{*}{\textbf{Baseline}} & \multicolumn{2}{c}{\textbf{NRMSE($\downarrow$)}} & \multicolumn{2}{c}{\textbf{ED($\downarrow$)}} & \multicolumn{2}{c|}{\textbf{PMGK($\uparrow$)}} & \multicolumn{2}{c}{\textbf{ACC($\uparrow$)}} & \multicolumn{2}{c}{\textbf{FID($\uparrow$)}} & \multicolumn{2}{c}{\textbf{PWD($\downarrow$)}} \\ \cline{2-3} \cline{5-16} 
                                   & Random                       & Worst                        &                                    & Random           & Worst           & Random         & Worst          & Random           & Worst           & Random          & Worst          & Random          & Worst          & Random          & Worst          \\ \hline
\multirow{3}{*}{\textbf{Single}}   & \multirow{3}{*}{79.66/96.10} & \multirow{3}{*}{79.63/96.09} & Rand.                              & 11.51            & 6.29            & 127.87         & 120.86         & 81.09            & 67.11           & 12.20           & 13.85          & 13.51           & 14.40          & 16.19           & 14.64          \\
                                   &                              &                              & FewE                               & 5.68             & 3.10            & 115.91         & 96.78          & 82.98            & 67.97           & 49.40           & 65.56          & 47.62           & 58.30          & 11.27           & 9.25           \\
                                   &                              &  & GLG-GU  &\textbf{3.74} &\textbf{2.13} &143.00 &149.78 &\textbf{84.04} &67.40 &0.36 &2.77 &1.43 &1.14 &18.99 &18.98   \\
                                   &                              &                              & Toxin                              & 4.76             & 2.49            & \textbf{87.37}          & \textbf{80.08}          & 81.56            & \textbf{68.30}           & \textcolor{blue}{\textbf{72.74}}           & \textcolor{blue}{\textbf{82.07}}          & \textcolor{blue}{\textbf{70.95}}           & \textcolor{blue}{\textbf{73.64}}          & \textcolor{blue}{\textbf{8.28}}            & \textcolor{blue}{\textbf{6.97}}           \\ \hline
\multirow{3}{*}{\textbf{Multiple}} & \multirow{3}{*}{79.86/96.09} & \multirow{3}{*}{79.78/96.10} & Rand.                              & 2.34             & 1.71            & 126.49         & 135.71         & 65.10            & 57.04           & 27.88           & 9.94           & 21.81           & 7.57           & 14.91           & 17.83          \\
                                   &                              &                              & FewE                               & 2.35             & 1.72            & 111.55         & 114.68         & \textbf{69.47}            & 59.83           & 41.92           & 38.55          & 33.20           & 33.63          & 13.22           & 12.76          \\
                                   &                              &  & GLG-GU  &\textbf{1.45} &1.12 &110.77 &124.04 &65.36 &\textbf{67.99} &22.33 &7.89 &17.82 &6.51 &15.99 &18.22   \\
                                   &                              &                              & Toxin                              & 1.53             & \textbf{1.07}            & \textbf{90.04}          & \textcolor{blue}{\textbf{71.66}}          & 67.70            & 59.93           & \textcolor{blue}{\textbf{59.41}}           & \textcolor{blue}{\textbf{69.60}}          & \textcolor{blue}{\textbf{47.88}}           & \textcolor{blue}{\textbf{60.11}}          & \textbf{11.56}           & \textcolor{blue}{\textbf{9.63}}           \\ \hline
\end{tabular}}
\end{table*}

\subsection{Ripple Effects - Deletion Size}
\label{ripple_size}

\begin{figure}[h]
    \centering
    \includegraphics[width=0.99\linewidth]{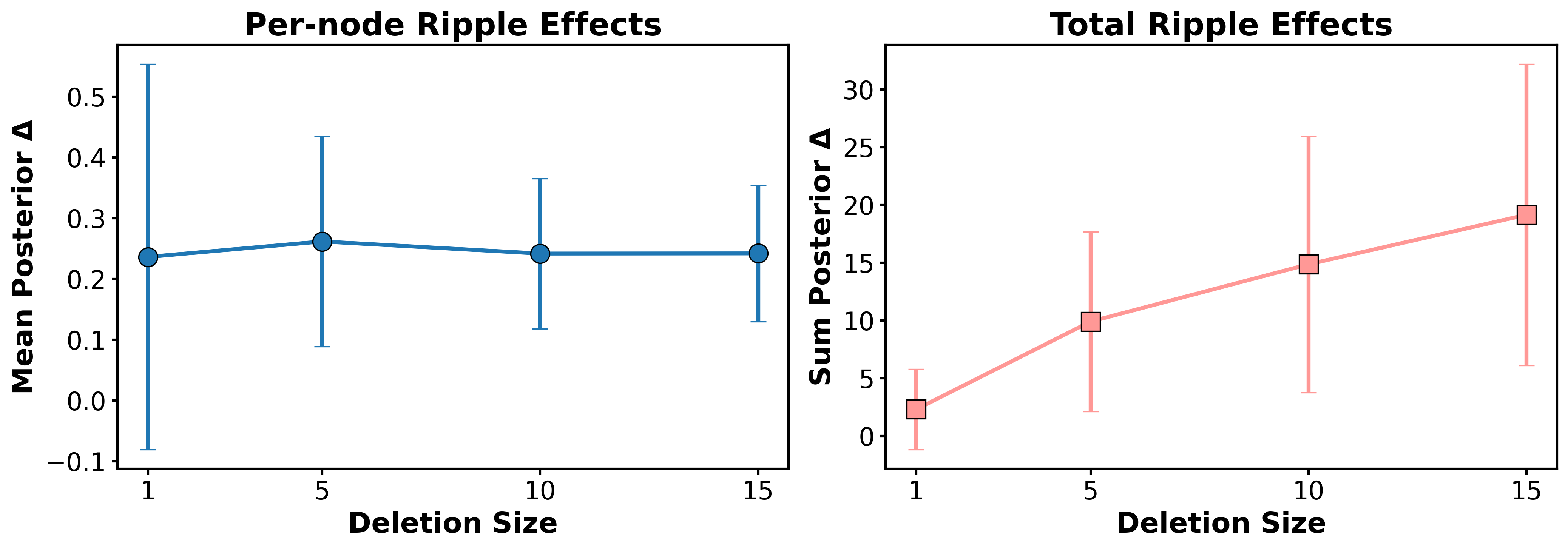}
    \caption{Impact of Deletion Size in Ripple Effects}
    \label{size_ripple_viz}
\end{figure}

As shown in Figure \ref{size_ripple_viz}, the per-node mean posterior change remains consistent across deletion sizes, indicating that the perturbation magnitude of each affected node is invariant to how many nodes are removed. In contrast, the total effects grow near-linearly with the deletion size. This finding suggests increasing the deletion size does not inherently amplify per-node vulnerability, but it broadens the attack surface by accumulating localized perturbations across the graph, potentially facilitating reconstruction of unlearned graphs.

\bibliographystyle{plainurl}
\bibliography{Reference}

\end{document}